\documentclass{article}

\usepackage[preprint]{neurips_2026}

\usepackage[utf8]{inputenc} 
\usepackage[T1]{fontenc}    
\usepackage{hyperref}       
\usepackage{url}            
\usepackage{booktabs}       
\usepackage{amsfonts}       
\usepackage{nicefrac}       
\usepackage{microtype}      
\usepackage{xcolor}         

\usepackage{caption}
\usepackage{graphicx}
\usepackage{booktabs}

\usepackage{bm}
\usepackage{algorithm}
\usepackage{algorithmic}
\usepackage{color, xcolor}
\usepackage{colortbl}
\definecolor{Gray}{RGB}{245, 236, 236}

\usepackage{pifont}
\newcommand{\cmark}{\ding{51}}%

\usepackage{amsmath,amssymb} 

\usepackage{booktabs} 
\usepackage{amssymb}
\usepackage{multirow}
\usepackage{array}
\usepackage{chngcntr} 

\usepackage[most]{tcolorbox}
\usepackage{xcolor}

\title{GIFT: Representation Geometry Matters for Single-Domain Generalized Object Detection}

\author{
  \normalfont
  \textbf{Yin Zhang}\textsuperscript{1,2}\quad
  \textbf{Yaoyue Zheng}\textsuperscript{2,3}\thanks{Corresponding author.}\\
  \textbf{Yongqiang Zhang}\textsuperscript{4}\quad
  \textbf{Bogdan Raducanu}\textsuperscript{2}\quad
  \textbf{Dan Liu}\textsuperscript{1}\\[0.5em]
  \textsuperscript{1}School of Instrument Science and Engineering, Harbin Institute of Technology\\
  \textsuperscript{2}Computer Vision Center, Universitat Autònoma de Barcelona\\
  \textsuperscript{3}Institute of Artificial Intelligence and Robotics, Xi'an Jiaotong University\\
  \textsuperscript{4}College of Computer Science, Inner Mongolia University\\
  \textit{
  \{yin.zhang.hit, yaoyz105\}@gmail.com}\\
  \textit{zhangyongqiang@imu.edu.cn, bogdan@cvc.uab.es, liudan@hit.edu.cn}
}

\begin{document}

\maketitle
\setcounter{footnote}{0}

\begin{abstract}
  Single-Domain Generalized Object Detection (S-DGOD) aims to generalize from a single source domain to unseen domains under distribution shifts. Existing efforts mainly focus on simulating unseen domains or suppressing domain-specific factors. However, they largely overlook the intrinsic generalization capability embedded in Vision Foundation Models (VFMs). In this work, we show that unconstrained fine-tuning disrupts the geometric structure of pre-trained representations, which encodes domain-invariant semantics. Motivated by this, we propose \textbf{G}eometric-\textbf{I}nvariance \textbf{F}ine-\textbf{T}uning (\textbf{GIFT}), a geometric-aware fine-tuning method for VFMs. Specifically, we introduce a series of learnable Householder reflections to the pre-trained weights to encourage geometry-consistent adaptation under domain shift. In addition, we propose a lightweight channel-wise scaling to facilitate effective task-specific adaptation. Extensive experiments demonstrate that GIFT outperforms state-of-the-art (SOTA) approaches on two S-DGOD benchmarks. Notably, by introducing fewer than 1\% additional trainable parameters into the frozen backbone, the proposed method achieves improvements of +7.9\% and +2.9\% in mPC over the previous SOTA on the Cityscapes-C and Diverse Weather Dataset, respectively. This finding highlights that preserving representation geometry is crucial for robust domain generalization, establishing GIFT as a principled and efficient solution to domain shift. The code is available at \url{https://github.com/zhangyin1996/GIFT}.
\end{abstract}

\section{Introduction}
\label{sec:intro}

Single-Domain Generalized Object Detection (S-DGOD)~\cite{CDSD-DWD} aims to learn from a single source domain while generalizing to diverse unseen target domains.
Traditional CNN-based methods primarily attempt to simulate domain diversity within a single source domain~\cite{physaug} or explicitly disentangle domain-invariant and domain-specific features~\cite{CDSD-DWD}. Despite their effectiveness, these methods are generally built on CNN backbones such as ResNet~\cite{resnet} and pre-trained on medium-scale datasets (e.g., ImageNet~\cite{imagenet}). Such constraints may limit representation capacity and hinder generalization to diverse unseen domains.

\begin{figure}[t]
\centering
\includegraphics[width=\textwidth]{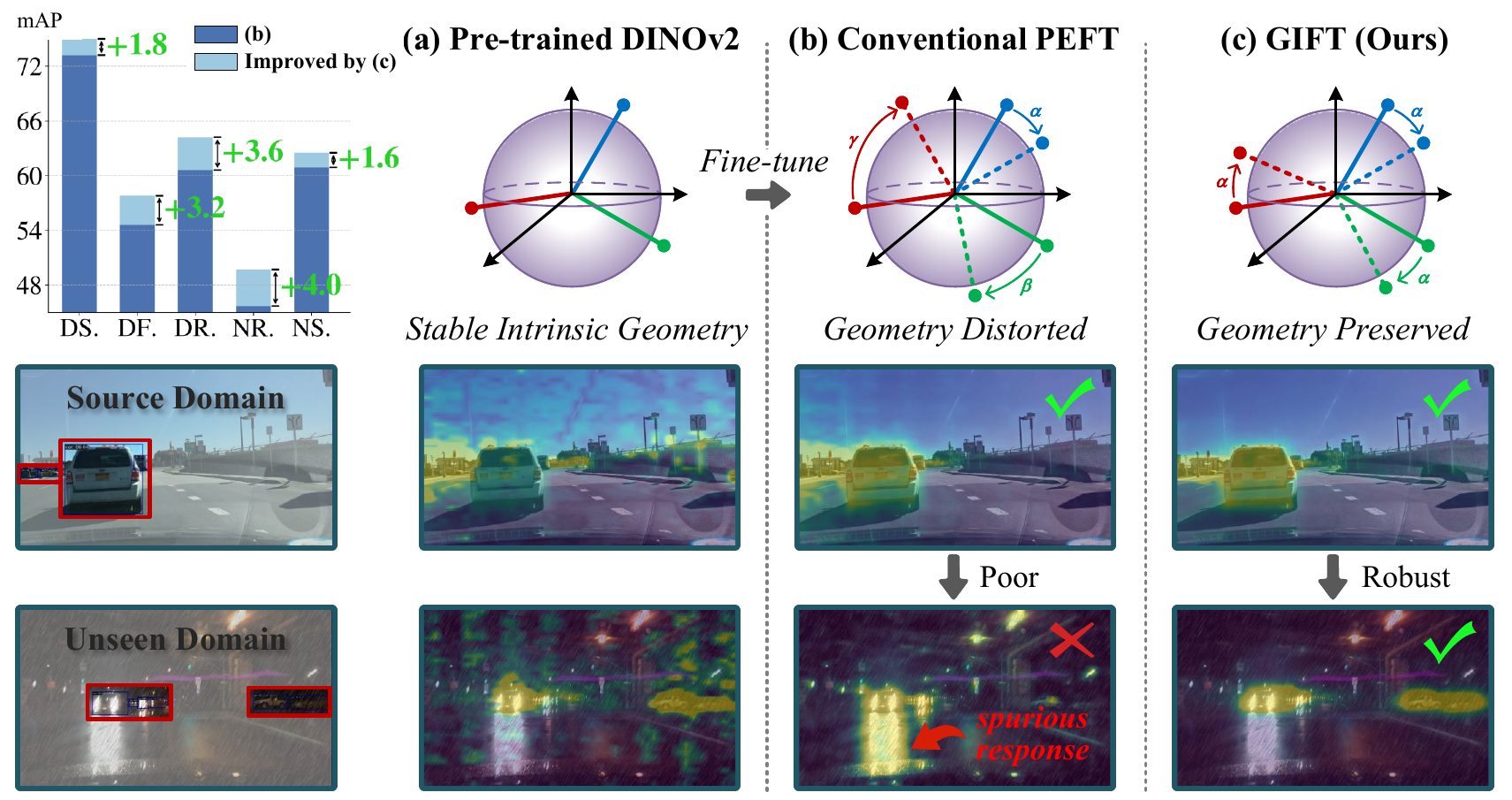}
\caption{\textbf{(a)} The pre-trained VFM exhibits an intrinsically stable geometry, while \textbf{(b)} unconstrained fine-tuning~\cite{VPT} often disrupts this structure, degrading generalization. In contrast, \textbf{(c)} by preserving geometric structure during fine-tuning, GIFT effectively improves generalization across unseen domains (DF., DR., NR., and NS.) in the DWD dataset~\cite{CDSD-DWD}. More results are provided in Section~\ref{sec:visual_results}.
}
\label{fig1_rotation_motivation}
\end{figure}

Recently, leveraging powerful Vision Foundation Models (VFMs), such as DINO~\cite{dinov1, dinov2, dinov3}, EVA~\cite{eva01, eva02}, and SAM~\cite{SAM1, SAM2, SAM3}, has become an effective strategy to overcome performance bottlenecks. With large-scale pre-training and strong representation capacity, these models have significantly advanced various vision tasks~\cite{abou2024parameter, yun2025soma, zhang2025causal}.
A mainstream approach to exploiting these VFMs is Parameter-Efficient Fine-Tuning (PEFT)~\cite{han2024parameter}, which adapts the pre-trained model by updating only a small subset of parameters through lightweight modules such as prompts~\cite{jiang2024t}, adapters~\cite{adaptformer}, or low-rank adaptations (LoRA)~\cite{lora}.
However, existing PEFT-based methods mainly focus on effectively adapting VFMs to downstream tasks. In the context of S-DGOD, the lack of multi-source data makes models prone to overfitting on source-specific statistics. Moreover, current PEFT-based S-DGOD methods largely follow conventional strategies from CNN-based methods, often overlooking the intrinsic generalization ability of VFMs, which is especially important under limited source diversity.

\paragraph{Motivation.}
We perform two experiments to show that unconstrained fine-tuning can disrupt the geometric structure of pre-trained representations, leading to degraded generalization on unseen target domains. By explicitly preserving this structure, our method improves cross-domain generalization.

\textit{\textbf{Experiment 1: Attention Map Visualization.}} 
As shown in Fig.~\ref{fig1_rotation_motivation}, 
we produce the mean CLS attention maps of the last layer~\cite{darcet2023vision-registers, jiang2025vision, shi2026vision} in DINOv2 for both the source and target domains of the DWD dataset~\cite{CDSD-DWD}, generated by \textbf{(a)} the pre-trained DINOv2, \textbf{(b)} an existing PEFT method~\cite{VPT}, and \textbf{(c)} our fine-tuning method. 
Compared with the existing PEFT method, our method produces attention maps that more clearly highlight foreground objects in both the source and target domains. These observations suggest that preserving the geometric structure of pre-trained representations helps the model maintain consistent attention to foreground objects across different domains, which is beneficial for single-domain generalization.

\begin{figure}[t]
\centering

\begin{minipage}[t]{0.56\textwidth}
    \centering
    \vspace{0pt}
    \includegraphics[width=\textwidth]{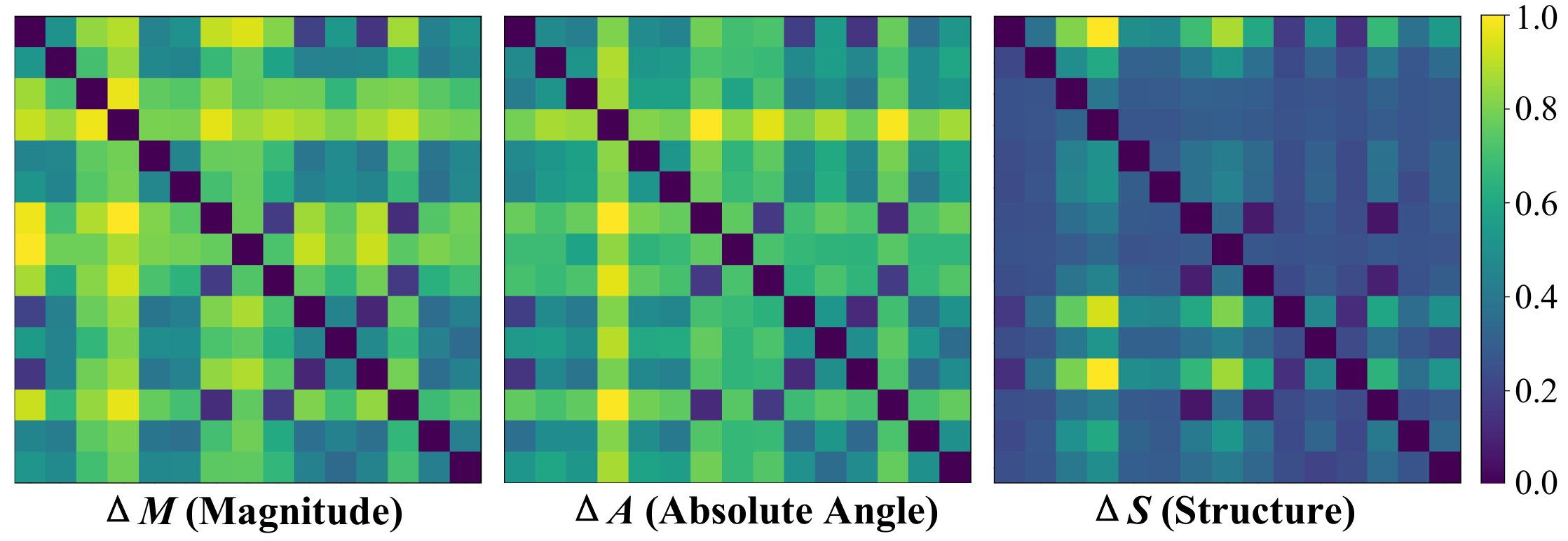}
    \caption{Geometric statistical analysis of DINOv2. 
    }
    \label{fig_vis_dinov2_dg_supp}
\end{minipage}
\hfill
\begin{minipage}[t]{0.41\textwidth}
    \centering
    \captionof{table}{The differences before and after fine-tuning for each PEFT method. `$\downarrow$' indicates that lower values are better.}
    \label{tab:before_and_after}
    \scriptsize
    \begin{tabular}{l|ccc}
    \toprule
    \textbf{Methods} 
    & $D_M (\downarrow)$ 
    & $D_A (\downarrow)$ 
    & $D_S (\downarrow)$ \\
    \midrule
    LoRA~\cite{lora}   & \textbf{0.18} & 0.085 & 0.11 \\
    Cauvis~\cite{Cauvis} & 0.25 & 0.13 & 0.19 \\
    \midrule
    Ours               & 0.20 & \textbf{0.051} & \textbf{0.081} \\
    \bottomrule
    \end{tabular}
\end{minipage}

\end{figure}

\textit{\textbf{Experiment 2: Geometric Statistical Analysis.}} 
We conduct a geometric statistical analysis to further support our motivation. First, for each input from Cityscapes-C~\cite{Cityscapes-C}, we feed the image into the DINOv2~\cite{dinov2} backbone and obtain the CLS token from the final transformer layer as the global representation. Then, inspired by PAID~\cite{wang2025paid}, we analyze the geometric properties of the representation space from three perspectives: Magnitude ($M$), Absolute Angle ($A$), and Structural Correlation ($S$). Specifically, given the feature matrix $F \in \mathbb{R}^{N \times D}$, where $N$ represents the total number of image samples in the batch, and $D$ denotes the feature dimension. The Magnitude ($M$) of each feature channel $d$ is defined as: 
\begin{equation}
M = \left\| F_{:,\: d} \right\|_2
\label{eq:magnitude}
\end{equation}
For absolute angle, we first normalize each channel vector: $\tilde{F}_{:,\:d} = F_{:,\:d}/{\|F_{:,\:d}\|_2}$, and we compute the global direction $\tau=\frac{1}{D}\sum_{d=1}^{D}\tilde{F}_{:,\: d}$, 
the Absolute Angle ($A$) of each feature channel $d$ is:
\begin{equation}
A = \arccos(\tau^{\top}\tilde{F}_{:,\:d})
\label{eq:angle}
\end{equation}
where $\tau$ is $L_2$-normalized before computing the cosine angle. The Structural Correlation ($S$) between channels is defined as:
\begin{equation}
S = \tilde{F}^{\top}\tilde{F}
\label{eq:structure}
\end{equation}
To quantify the representation differences between corruption domains, we compute pairwise distances of the above geometric descriptors. For two domains $i$ and $j$, we define three metrics: Magnitude difference $\Delta{M}$, Absolute Angle difference $\Delta{A}$, and Structural difference $\Delta{S}$. For fair visualization, all three geometric differences are independently normalized to [0,1] in Fig.~\ref{fig_vis_dinov2_dg_supp}.

In addition, we compare the geometric statistics of different PEFT methods before and after fine-tuning to provide stronger evidence. Specifically, using Eqs.~(\ref{eq:magnitude})-(\ref{eq:structure}), we compute the Magnitude ($M$), Absolute Angle ($A$), and Structural Correlation ($S$) of the representations before and after fine-tuning for each PEFT method, denoted as $(M_{\mathrm{before}}, A_{\mathrm{before}}, S_{\mathrm{before}})$ and $(M_{\mathrm{after}}, A_{\mathrm{after}}, S_{\mathrm{after}})$, respectively. Then, we quantify the relative changes in magnitude, absolute angle, and structural correlation before and after fine-tuning using normalized $L_2$ distances for $M$ and $A$, and the normalized Frobenius distance for $S$, denoted as $D_M$, $D_A$, and $D_S$, respectively.



As shown in Fig.~\ref{fig_vis_dinov2_dg_supp}, the Magnitude ($\Delta M$) and Absolute Angle difference ($\Delta A$) exhibit substantial variations across corruption domains, whereas the Structural difference ($\Delta S$) shows a more consistent pattern across different corruption pairs. Moreover, the results in Table~\ref{tab:before_and_after} demonstrate that our method preserves representation geometry better than other PEFT methods after fine-tuning. This observation suggests that the structural relationships of DINOv2 representations are relatively stable under corruption shifts. However, the unconstrained fine-tuning of conventional adapter-based PEFT methods may disrupt this intrinsic geometric structure. Therefore, this motivates us to preserve the intrinsic geometric structure during fine-tuning.

\paragraph{Contribution.}
Building on these insights, we propose \textbf{G}eometric-\textbf{I}nvariance \textbf{F}ine-\textbf{T}uning (\textbf{GIFT}) for S-DGOD, which incorporates two key innovations. 
First, we preserve the geometric structure of pre-trained representations by introducing a series of Householder reflections~\cite{yuan2024bridging}, which ensure that the parameter space is transformed without altering its geometry.
Second, we introduce a lightweight channel-wise scaling to the transformed weights to adapt task-specific features. Built on the geometric-preserving fine-tuning, this simple scaling yields noticeable gains in downstream object detection tasks, further highlighting the importance of maintaining the pre-trained geometric structure during fine-tuning.
In summary, the main contributions of this paper are as follows:

\begin{itemize}

\item We are the first to investigate fine-tuning pre-trained VFMs for S-DGOD from a geometric perspective, revealing that unconstrained updates can distort representations and impair cross-domain generalization.

\item We propose \textbf{GIFT}, a geometric-invariance fine-tuning method that preserves the intrinsic geometry of pre-trained representations, improving cross-domain generalization and enabling effective task-specific adaptation.


\item Extensive experiments on two S-DGOD benchmarks demonstrate the effectiveness of GIFT, achieving +7.9\% and +2.9\% mPC improvements over the previous SOTA on Cityscapes-C and Diverse Weather Dataset, respectively.

\end{itemize}



\section{Related Work}

\paragraph{Single-Domain Generalized Object Detection (S-DGOD).}

Most S-DGOD methods improve generalization either by simulating domain diversity through augmentation/diversification~\cite{physaug, danish2024improving, OA-DG}, or by learning domain-invariant representations~\cite{lin2021domain, CDSD-DWD, UFR}. More recently, CLIP-based methods~\cite{CLIP, ClipGap, xu2025boosting, yu2026semclip} have shown promising results, but most methods still rely on full-parameter fine-tuning.

Leveraging the power of large pre-trained Vision Foundation Models (VFMs), PEFT~\cite{han2024parameter, xu2026parameter} has emerged as a promising direction for vision tasks~\cite{abou2024parameter, szatkowski2024improving, zheng2026revisiting}. 
However, S-DGOD under this paradigm remains largely underexplored, with only a few recent attempts~\cite{Cauvis, yun2025soma, jiang2025diffusion}. Specifically, Cauvis~\cite{Cauvis} adopts DINOv2~\cite{dinov2} as a backbone and employs causal intervention to suppress spurious domain-specific correlations. Similarly, SoMA~\cite{yun2025soma} applies Singular Value Decomposition to separate domain-specific dominant components from domain-invariant semantic components. 
However, these methods overlook the intrinsic generalization ability of VFMs.
\textit{In this work, we study PEFT-based S-DGOD from the perspective of the geometric structure of pre-trained representations. We reveal that maintaining a stable representation geometry is more critical for generalization than explicitly removing domain-specific features.}

\paragraph{Geometric Invariance in Representation Learning.}

Geometric invariance is crucial for representation learning. Early works such as SphereFace~\cite{liu2017sphereface} and ArcFace~\cite{deng2019arcface} established this foundation by projecting features onto a hypersphere and optimizing angular margins, showing that angular distributions are more discriminative and robust. 
In domain-shift learning, AIDGN~\cite{jin2022domain} and PAID~\cite{wang2025paid} interpret angular structures as domain-invariant semantic cues. However, AIDGN necessitates inter-domain contrastive signals to identify invariant components, whereas PAID relies on access to target-domain data for adaptation. 
Recently, Householder Transformations~\cite{yuan2024bridging, dong2024efficient} have been proposed to preserve the geometric structure of the learnable feature space through orthogonal adaptations. However, they aim to improve adaptation efficiency rather than enhance cross-domain generalization.
\textit{Building on these insights, this work
leverages geometric-preserving fine-tuning to tackle S-DGOD, enabling robust localization and recognition in unseen target domains.
}


\section{Proposed Method}

\subsection{Preliminaries}
\label{subsec:preliminaries}


\paragraph{Task Formulation.} 
\label{subsubsec:task}

Let $\mathcal{D}_s = \{(x_i, y_i)\}_{i=1}^{N_s}$ denote the source domain dataset, sampled from the distribution $(x, y) \sim P_s$. Our goal is to learn a detector that generalizes to any unseen target domain $\mathcal{D}_t \sim P_t$, where $P_s \neq P_t$. 

In this work, we initialize the detector with a pre-trained model parameterized by $\bm{W}_0$ and fine-tune it on the source domain dataset $\mathcal{D}_s$. The training objective is formulated as:
\begin{equation}
    \bm{W}_s = 
    \arg\min_{\bm{W}} 
    \mathbb{E}_{(x, y) \sim P_s}
    \big[ \mathcal{L}(f_{\bm{W}}(x), y) \big]
\end{equation}
where $\bm{W}_s$ is initialized from $\bm{W}_0$. In the S-DGOD setting, only $\mathcal{D}_s$ is accessible during training, while target domains remain unseen. Our objective is to learn source-optimal parameters $\bm{W}_s$ that generalize well to unseen target distributions $P_t$. Formally, we expect the target risk
\begin{equation}
    \mathcal{R}_t(\bm{W}_s) 
    = \mathbb{E}_{(x,y)\sim P_t}
    \big[ \mathcal{L}(f_{\bm{W}_s}(x), y) \big]
\end{equation}
to remain low for $P_t \neq P_s$.

\paragraph{Householder Reflection.}
\label{subsubsec:householder}

To mitigate structural distortion, we adopt the Householder reflection~\cite{yuan2024bridging} as a geometry-preserving adaptation mechanism. A Householder reflection refers to a linear transformation that performs a reflection with respect to a hyperplane defined by a unit vector $\bm{v} \in \mathbb{R}^{d}$, referred to as the Householder vector. The corresponding Householder matrix $\bm{H} \in \mathbb{R}^{d \times d}$ is constructed as follows:
\begin{equation}
    \bm{H} = \bm{I} - 2 \bm{v} \bm{v}^{\top} \quad \text{subject to } \|\bm{v}\|_2 = 1
\end{equation}
where $\bm{I}$ denotes the identity matrix and superscript $\top$ indicates the transpose of a matrix. A key geometric property of $\bm{H}$ is its \textbf{orthogonality} ($\bm{H}^{\top}\bm{H} = \bm{I}$), which ensures that the transformation is an isometry. For any two feature vectors $\bm{x}_1, \bm{x}_2$, the transformation preserves their inner product:
\begin{equation}
    \langle \bm{H}\bm{x}_1, \bm{H}\bm{x}_2 \rangle = (\bm{H}\bm{x}_1)^{\top} (\bm{H}\bm{x}_2) = \bm{x}_1^{\top} \bm{H}^{\top} \bm{H} \bm{x}_2 = \bm{x}_1^{\top} \bm{x}_2
\end{equation}

This property forms the theoretical basis of our method, ensuring that geometric relationships among pre-trained feature representations remain stable during fine-tuning.


\begin{figure}[t]
\centering
\includegraphics[width=\textwidth]{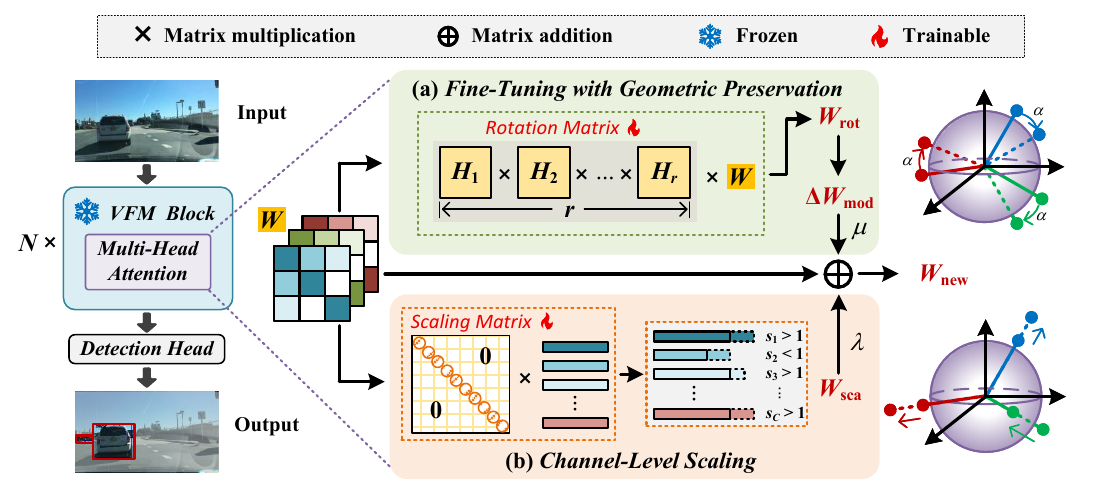}
\caption{An overview of our proposed method. It is composed of two key components: \textbf{(a)} Fine-Tuning with Geometric Preservation (Section~\ref{subsec:method1}) and \textbf{(b)} Channel-Level Scaling (Section~\ref{subsec:method2}). Notably, we optimize only the multi-head attention parameters within each block. The updated weights, $\bm{W}_{\text{new}}$, are obtained by adding the scaled residual and channel-wise adjustment to the original weights $\bm{W}$.}
\label{fig_pipeline}
\end{figure}

\subsection{Fine-Tuning with Geometric Preservation}
\label{subsec:method1}

As shown in Fig.~\ref{fig_pipeline}, for a frozen VFM with $N$ Transformer blocks, we propose a geometric-preserving fine-tuning strategy that adapts the model while maintaining the intrinsic geometry of the pre-trained representation space.
Instead of directly updating the backbone weights, we introduce a learnable rotation matrix $\bm{R}$, defined as the product of $r$ Householder Reflection (HR) matrices, and integrate it into the frozen weights of the \textit{Q}, \textit{K}, \textit{V}, and \textit{Output} projection layers. The transformed weights $\bm{W}_{\text{rot}} $ can be represented as:  
\begin{equation}
    \bm{W}_{\text{rot}} 
    = \bm{R} \bm{W}
    = \left( \prod_{i=1}^{r} \bm{H}_i \right) \bm{W}
    = \left( \prod_{i=1}^{r} (\bm{I} - 2 \bm{v}_i \bm{v}_i^\top) \right) \bm{W}
\end{equation}
where $\bm{H}_i = \bm{I}-2\bm{v}_i \bm{v}_i^\top$ denotes the $i$-th HR matrix and $\bm{v}_i$ is a learnable unit vector. $\bm{W}$ denotes the original frozen projection weights. 
Since each HR matrix $\bm{H}_i$ is orthogonal, the product of orthogonal matrices remains orthogonal, thus the transformation $\bm{R}$ also satisfies $\bm{R}^{\top} \bm{R} = \bm{I}$. Consequently, for any pair of weight vectors $\bm{w}_1$ and $\bm{w}_2$, their inner product is preserved:
\begin{equation}
    \langle \bm{R} \bm{w}_1, \bm{R} \bm{w}_2 \rangle
    = \langle \bm{w}_1, \bm{w}_2 \rangle
\end{equation}
In particular, the $\ell_2$ norm and pairwise distances remain invariant:
\begin{equation}
    \|\bm{R} \bm{w}\|_2 = \|\bm{w}\|_2, 
    \qquad
    \|\bm{R} \bm{w}_1 - \bm{R} \bm{w}_2\|_2 
    = \|\bm{w}_1 - \bm{w}_2\|_2
\end{equation}
These properties ensure that $\bm{R}$ rotates the parameter space without altering its geometry, thereby preserving the structural stability of the pre-trained model and mitigating distortions from direct weight updates.

In practice, directly applying $\bm{R}$ may induce large changes in the pre-trained weights. To ensure stable optimization, we employ a residual update mechanism. Specifically, we first compute the difference between the rotated and original weights, and then modulate this residual using a learnable gate vector $\bm{g}\in \mathbb{R}^{D_{out}}$, initialized to all ones so that all output channels contribute equally at the start of training:
\begin{equation}
    \Delta \bm{W}_{\text{mod}} = \bm{g} \odot \Delta \bm{W} = \bm{g} \odot (\bm{W}_{\text{rot}} - \bm{W}) = \bm{g} \odot (\bm{R} \bm{W} - \bm{W})
\label{eq:residual_update}
\end{equation}

where $\odot$ denotes element-wise multiplication across columns of $\Delta \bm{W}$. The gate $\bm{g}$ allows selective adjustment of the residual along different directions, effectively controlling which components of the weight space are rotated.

\subsection{Channel-Level Scaling for Stable Feature Selection}
\label{subsec:method2}

Preserving the pretrained semantic structure is necessary but not sufficient for single-domain generalization. To identify task-specific features that are consistent under source-domain variations, we introduce a learnable channel-wise scaling matrix $\bm{S}$, initialized to zeros:
\begin{equation}
    \bm{W}_{\text{sca}} = \bm{S} \bm{W}, \quad \bm{S} = \mathrm{diag}(s_1, \dots, s_C)
\end{equation}
where $\bm{W}_{\text{sca}}$ denotes the re-scaled weights, and each diagonal element of $\bm{S}$ modulates the importance of a single feature channel.

To measure feature stability, we adopt gradient consistency as a proxy. Let $f_{i}(x)$ denote the output of the $i$-th feature channel. The gradient of the detection loss $\mathcal{L}$ with respect to the scaling $s_{i}$ is:
\begin{equation}
    \mathcal{G}_i = \frac{\partial \mathcal{L}}{\partial f^{'}_{i}} f_{i}(x), \quad f_{i}^{'}(x) = s_i f_{i}(x)
\end{equation}

The expected gradient over the source domain can be viewed as an indicator of how consistently channel $i$ contributes across intra-domain variations. Channels that consistently contribute to loss reduction exhibit non-zero expected gradients, while spurious channels tend to produce gradients with high variance but near-zero mean. During stochastic optimization, gradients are accumulated over iterations:
\begin{equation}
    s_{i}^{T} = s_{i}^{0} - \eta \sum_{t=1}^{T} \mathcal{G}_{i}^{t}
\end{equation}
where $\eta$ denotes the learning rate. Under SGD optimization, channels with consistently aligned gradients tend to accumulate larger updates over training iterations, while unstable channels with highly varying gradients are less consistently reinforced.

The final updated weights $\bm{W}_{\text{new}}$ are obtained by adding the scaled residual to the original weights, together with a diagonal scaling branch to allow per-channel adjustment:
\begin{equation}
     \bm{W}_{\text{new}}
    = \bm{W} + \mu \, \Delta \bm{W}_{\text{mod}} 
    + \lambda \, \bm{W}_{\text{sca}}
\end{equation}
where the balancing coefficients $\mu$ and $\lambda$ are implemented as learnable scalar parameters.


Importantly, the orthogonal rotation branch primarily preserves relational geometric structure, such as pairwise angular relationships and relative orientations between pretrained weight vectors.
In contrast, the channel-level scaling components intentionally modify feature magnitudes and channel importance to enable task-specific adaptation. Therefore, the proposed method should be viewed as preserving stable structural relations while allowing controlled feature reweighting, rather than enforcing strict isometry during adaptation.



\section{Experiments}
\label{sec:experiments}


\subsection{Benchmarks}
\label{subsec:benchmarks}

\paragraph{Datasets and Evaluation Metrics.}

Following existing S-DGOD methods~\cite{AutoAug, OA-DG, Cauvis}, we evaluate our method on two benchmarks: the Diverse Weather Dataset (DWD)~\cite{CDSD-DWD} and Cityscapes-C~\cite{Cityscapes-C}.
\textbf{DWD} 
comprises five scenarios: Daytime-Sunny (DS.), Daytime-Foggy (DF.), Dusk-Rainy (DR.), Night-Rainy (NR.), and Night-Sunny (NS.).
\textbf{Cityscapes-C} is a benchmark for evaluating detection robustness under common corruptions. It generates 15 corruption types with 5 severity levels. 


For evaluation, we report mean Average Precision (mAP$_{50}$) to measure detection performance and mean Performance under Corruption (mPC) to evaluate robustness. For DWD~\cite{CDSD-DWD}, mPC is computed as the average mAP over four weather conditions, while for Cityscapes-C~\cite{Cityscapes-C}, it averages mAP$_{50}$ across all corruption types and severity levels.

\paragraph{Implementation Details.}

We use DINOv2-Large~\cite{dinov2} as the default VFM, while results with other VFMs are reported in Appendix~\ref{appendix:VFMs}. For DWD, we employ Faster R-CNN~\cite{fasterrcnn} and DINO~\cite{dino-detr-detector} (a DETR-based detector) as detection heads, respectively. For Cityscapes-C, Faster R-CNN is used as the detection head. All models are trained for 12 epochs using the AdamW optimizer with $\beta_1 = 0.9$ and $\beta_2 = 0.999$, a learning rate of $1 \times 10^{-4}$, and a batch size of 2. By default, the number of HR matrices is set to $r = 6$, and only the last $n=16$ Transformer blocks are fine-tuned. All experiments are implemented in \textit{MMDetection}~\cite{mmdetection} and conducted on a single RTX 3090 GPU.

\subsection{Comparison with State-of-the-art S-DGOD Methods}

\begin{table}[t]
\centering
\caption{Comparison with SOTA methods on Cityscapes-C~\cite{Cityscapes-C}. mPC denotes the average performance across 15 different corruption types. For fairness, we adopt Faster R-CNN~\cite{fasterrcnn} as the detection head. `Baseline' denotes a fully frozen DINOv2~\cite{dinov2} backbone with only the Faster R-CNN detection head.  
`$\dagger$' denotes unreported results re-implemented with the official code under the same settings.
}
\label{tab:cityscapes_c}
\resizebox{\textwidth}{!}{
\begin{tabular}{l | c | ccc | cccc | cccc | cccc | c}
\toprule
 & 
& \multicolumn{3}{c|}{Noise}
& \multicolumn{4}{c|}{Blur} 
& \multicolumn{4}{c|}{Weather}
& \multicolumn{4}{c|}{Digital}
 \\
\cmidrule(lr){3-5} \cmidrule(lr){6-9} \cmidrule(lr){10-13} \cmidrule(lr){14-17}
\textbf{Methods} &Norm. & Gau. & Sho. & Imp. & Def. & Gla. & Mot. & Zoo. & Sno. & Fro. & Fog & Bri. & Con. & Ela. & Pix. & JPE. & \textbf{mPC} \\
\midrule
\multicolumn{18}{c}{\textbf{Traditional S-DGOD}} \\
FRCNN~\cite{fasterrcnn} & 42.2 & 0.5 & 1.1 & 1.1 & 17.2 & 16.5 & 18.3 & 2.1 & 2.2 & 12.3 & 29.8 & 32.0 & 24.1 & 40.1  & 18.7 & 15.1 & 15.4 \\
AutoAug~\cite{AutoAug} & 42.4 & 0.9 & 1.6 & 0.9 & 16.8 & 14.4 & 18.9 & 2.0 & 1.9 & 16.0 & 32.9 & 35.2 & 26.3 & 39.4 & 17.9 & 11.6 & 15.8 \\
AugMix~\cite{Augmix} & 39.5 & 5.0 & 6.8 & 5.1 & 18.3 & 18.1 & 19.3 & 6.2 & 5.0 & 20.5 & 31.2 & 33.7 & 25.6 & 37.4 & 20.3 & 19.6 & 18.1 \\
Stylized~\cite{Stylized} & 36.3 & 4.8 & 6.8 & 4.3 & 19.5 & 18.7 & 18.5 & 2.7 & 3.5 & 17.0 & 30.5 & 31.9 & 22.7 & 33.9 & 22.6 & 20.8 & 17.2 \\
SupCL~\cite{SupCL} & 43.2 & 7.0 & 9.5 & 7.4 & 22.6 & 20.2 & 22.3 & 4.3 & 5.3 & 23.0 & 37.3 & 38.9 & 31.6 & 40.1 & 24.0 & 20.1 & 20.9 \\
FSCE~\cite{FSCE} & 43.1 & 7.4 & 10.2 & 8.2 & 23.3 & 20.3 & 21.5 & 4.8 & 5.6 & 23.6 & 37.1 & 38.0 & 31.9 & 40.0 & 23.2 & 20.4 & 21.0 \\
OA-DG~\cite{OA-DG} & 43.4 & 8.2 & 10.6 & 8.4 & 24.6 & 20.5 & 22.3 & 4.8 & 6.1 & 25.0 & 38.4 & 39.7 & 32.8 & 40.2 & 23.8 & 22.0 & 21.8 \\
\midrule
\multicolumn{18}{c}{\textbf{Parameter-Efficient Fine-Tuning S-DGOD}} \\
Baseline & 44.0 & 14.5 & 16.3 & 13.5 & 33.9 & 27.3 & 33.1 & 13.7 & 25.2 & 31.5 & 39.5 & 42.5 & 38.9 & 42.1 & 35.9 & 31.8 & 29.3 \\
SoMA$^\dagger$~\cite{yun2025soma} & 53.5 & 16.4 & 19.2 & \underline{15.3} & 39.4 & 31.4 & 37.7 & 15.0 & 28.2 & \underline{36.7} & 47.2 & 51.5 & 45.9 & 50.9 & 42.3 & 36.3 & 34.2 \\
Cauvis~\cite{Cauvis} &\underline{54.6} &\underline{16.8} &\underline{19.8} &15.2 & \underline{41.4} &\underline{34.0} &\underline{39.2} &\underline{15.8} &\underline{29.8} &\underline{36.7} &\underline{48.8} &\underline{53.0} &\underline{49.5} &\underline{52.0} &\underline{43.9} &\underline{38.8}  &\underline{35.6}  \\
GIFT (Ours) & \textbf{62.7} & \textbf{23.2} & \textbf{25.1} & \textbf{20.6} & \textbf{49.8} & \textbf{43.2} & \textbf{48.0} & \textbf{19.2} & \textbf{39.8} & \textbf{44.2} & \textbf{58.2} & \textbf{61.4} & \textbf{58.0} & \textbf{61.0} & \textbf{53.3} & \textbf{48.1} & \textbf{43.5} \\
\bottomrule
\end{tabular}
}
\end{table}

\paragraph{Performance on Cityscapes-C.}

We evaluate the robustness of GIFT under diverse synthetic corruptions. For Cityscapes-C~\cite{Cityscapes-C} dataset, the 15 corruption types are abbreviated as `Gau.', `Sho.', `Imp.', `Def.', `Gla.', `Mot.', `Zoo.', `Sno.', `Fro.', `Fog', `Bri.', `Con.', `Ela.', `Pix.', and `JPE.'. `Norm.' denotes the model’s performance evaluated on the source domain, Cityscapes-Clean~\cite{cordts2016cityscapes}.
From Table~\ref{tab:cityscapes_c}, GIFT demonstrates clear superiority over existing methods in terms of overall performance (mPC). Specifically, GIFT achieves an average mPC of 43.5\% across all 15 corruption types, outperforming the baseline (Frozen DINOv2~\cite{dinov2} with Faster R-CNN~\cite{fasterrcnn}, +14.2\%) and SOTA method Cauvis~\cite{Cauvis} (+7.9\%).

\begin{table}[t]
\centering
\scriptsize
\begin{minipage}[t]{0.5\textwidth}
\caption{Comparison with SOTA methods on DWD~\cite{CDSD-DWD}. mPC denotes the average performance across 4 different adverse weather conditions.
}
\label{tab:DWD}
\begin{tabular}{l | c| cccc | c}
\toprule
\textbf{Methods} & DS. & DF. & DR. & NR. & NS. & \textbf{mPC} \\
\midrule
\multicolumn{7}{c}{\textbf{Traditional S-DGOD}} \\
IN~\cite{IterNorm} & 43.9 & 28.5 & 22.8 & 12.6 & 29.6 & 23.4 \\
SW~\cite{SW} & 50.6 & 30.8 & 26.3 & 13.7 & 33.4 & 26.1 \\
ISW~\cite{ISW} & 51.3 & 31.8 & 25.9 & 14.1 & 33.2 & 26.3 \\
FRCNN~\cite{fasterrcnn} &56.2 & 33.5 & 28.0 & 14.2 & 35.8 & 27.9\\
CDSD~\cite{CDSD-DWD} & 56.1& 33.5 & 28.2 & 16.6 & 36.6 & 28.7 \\
OA-DG~\cite{OA-DG} &55.8 & 38.3 &33.9  &16.8  &38.0  &31.8 \\
ClipGap~\cite{ClipGap} & 52.5 & 38.5 & 32.3 & 18.7 & 36.9 & 31.6\\
G-NAS~\cite{G-nas} &56.1 & 36.4 & 35.1 & 17.4 & 45.0 & 33.5\\
PDD~\cite{PDDOC} &53.6 & 39.1 & 33.7 & 19.2 & 38.5 & 32.6\\
UFR~\cite{UFR} &58.6 & 39.6 & 33.2 & 19.2 & 40.8 & 33.2\\
\midrule
\multicolumn{7}{c}{\textbf{Parameter-Efficient Fine-Tuning S-DGOD}} \\
\multicolumn{7}{c}{\textit{Detection Head: Faster R-CNN}~\cite{fasterrcnn}} \\
SoMA~\cite{yun2025soma} & 49.3 & 38.2 & 37.9 & 24.5 & 41.9 & 35.6\\
Cauvis~\cite{Cauvis} &72.2 & \underline{52.3} & \underline{61.3} & \underline{44.1} & \underline{58.3} & \underline{54.1}\\
GIFT (Ours) & \textbf{73.9} & \textbf{54.9} & \textbf{64.4}  & \textbf{47.1}  & \textbf{61.7} & \textbf{57.0} \\
\midrule
\multicolumn{7}{c}{\textit{Detection Head: DETR-based DINO}~\cite{dino-detr-detector}} \\
SoMA~\cite{yun2025soma} & 69.4  & 51.0  & 59.3  & \underline{47.6}  &59.3  & 54.3\\
Cauvis~\cite{Cauvis} &\underline{73.7}  & \underline{56.5} & \underline{64.6} & \underline{47.6} & \underline{61.2}  & \underline{57.5}\\
GIFT (Ours)  & \textbf{76.0}  & \textbf{58.3} & \textbf{65.9} & \textbf{49.9} & \textbf{63.3}  & \textbf{59.4}\\
\bottomrule
\end{tabular}
\end{minipage}
\hfill
\begin{minipage}[t]{0.48\textwidth}
\centering
\scriptsize
\caption{Ablation study of geometric preservation (GP) and channel-wise scaling (CS) on DWD, using DINO~\cite{dino-detr-detector} as the detection head.}
\label{tab:ab_study_dwd}
\begin{tabular}{cc|cccc|c}
\toprule
\textbf{GP} & \textbf{CS}  & DF. & DR. & NR. & NS.  & \textbf{mPC} \\
\midrule
- & -   & 53.5 & 60.8 & 42.6 & 59.5  & 54.1\\
- & \cmark  & 57.1 & 62.9 & 46.8 & 60.7  & 56.9 \\
\cmark & -  & \underline{57.8} & \underline{64.2} & \underline{49.7} & \underline{62.5}  & \underline{58.6} \\ 
\cmark & \cmark  & \textbf{58.3} & \textbf{65.9} & \textbf{49.9} & \textbf{63.3}  & \textbf{59.4} \\
\bottomrule
\end{tabular}
\caption{Comparison of different PEFT methods on DINOv2 with DETR-based~\cite{dino-detr-detector} detector.}
\label{tab:different_PEFT}
\begin{tabular}{ l | c c c c | c}
\toprule
\textbf{Methods}  & DF. & DR. & NR. & NS. &\textbf{mPC}\\
\midrule
Frozen  & 53.5 & 60.8 & 42.6 & 59.5  & 54.1\\
\midrule
AdaFormer~\cite{adaptformer}  & 54.6 & 61.1 & 42.1 & 59.1  & 54.2\\
LoRA~\cite{lora} & 54.5  &61.2  &44.7  &59.3   &54.9  \\
VPT~\cite{VPT} & 54.6 & 60.6 & 45.7 & 60.9  & 55.5 \\
Rein~\cite{Rein} &55.0 & 62.4 & 45.2 &59.4 &55.5 \\
SPT~\cite{wang2024revisiting-SPT}  & 55.7 & 62.6 & 46.6 & 60.6  & 56.4 \\
CausalTune~\cite{zhang2025causal}  & \underline{57.1} & 63.8 & 47.2 & \underline{61.4}  & 57.4 \\
\midrule
SoMA~\cite{yun2025soma} & 51.0  & 59.3  & \underline{47.6}  &59.3  & 54.3\\
Cauvis~\cite{Cauvis}  & 56.5 & \underline{64.6} & \underline{47.6} & 61.2  & \underline{57.5}\\
GIFT (Ours)  & \textbf{58.3} & \textbf{65.9} & \textbf{49.9} & \textbf{63.3}   & \textbf{59.4}\\
\bottomrule
\end{tabular}
\end{minipage}
\end{table}

\paragraph{Performance on DWD.}

The results in Table~\ref{tab:DWD} demonstrate that our proposed method outperforms existing approaches across all adverse weather conditions and detector architectures.
In the traditional S-DGOD setting, prior methods achieve limited robustness, with the best method (G-NAS) reaching an mPC of 33.5\%. 
In contrast, parameter-efficient fine-tuning methods significantly improve performance, highlighting the effectiveness of adapting large pretrained models for S-DGOD. i) Under the Faster R-CNN detection head, our method GIFT achieves the best performance, reaching 57.0\% in mPC and outperforming SoMA (35.6\%) and Cauvis (54.1\%). Notably, GIFT improves performance across all weather conditions (DF., DR., NR., NS.), indicating its strong generalization ability. ii) When using the DETR-based DINO detector, GIFT further improves performance, achieving an mPC of 59.4\%, outperforming Cauvis (57.5\%) by +1.9\%.

\subsection{Ablation Studies and Analysis}
\label{subsec:ab_exp}

\paragraph{Effectiveness of Each Component.}

\begin{figure}[t]
\centering
\begin{minipage}[b]{0.32\textwidth}
    \centering
    \includegraphics[width=0.8\textwidth]{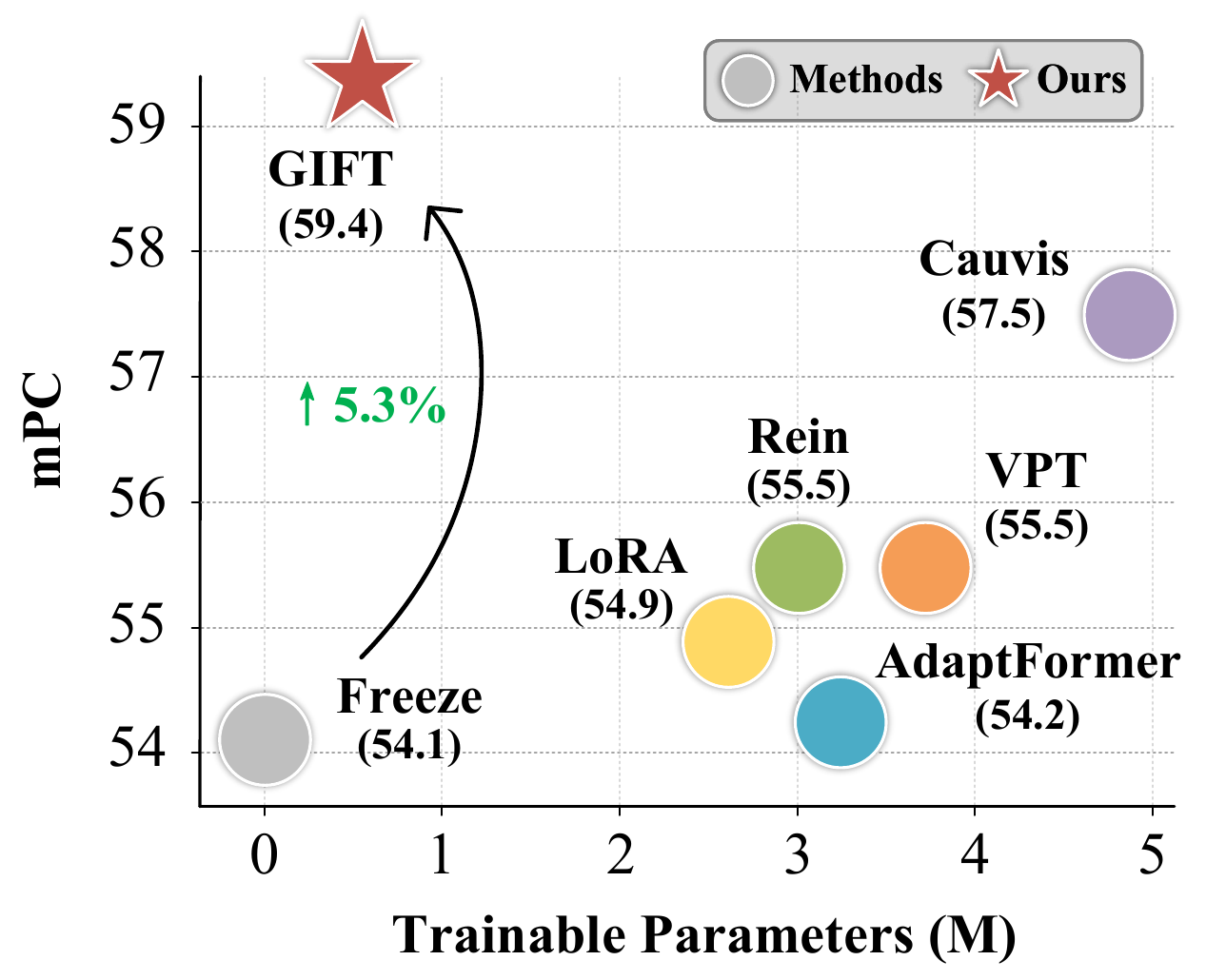}
    \caption{Trainable parameter efficiency of different methods based on DINOv2-Large~\cite{dinov2}. 
    }
    \label{fig:param_efficiency}
\end{minipage}
\hfill
\begin{minipage}[b]{0.32\textwidth}
    \centering
    \includegraphics[width=0.8\textwidth]{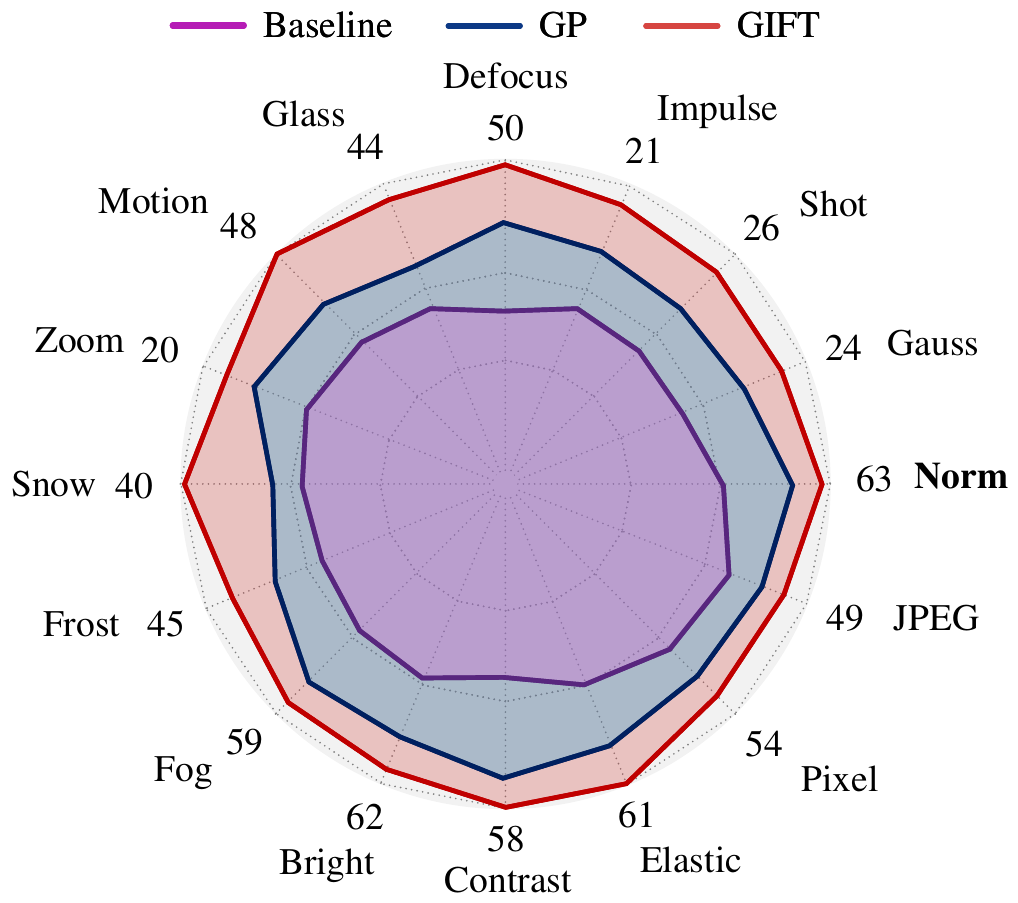}
    \caption{
    Ablation Studies on Cityscapes-C~\cite{Cityscapes-C} dataset.
    }
    \label{fig:ab_study_city}
\end{minipage}
\hfill
\begin{minipage}[b]{0.32\textwidth}
    \centering
    \includegraphics[width=\textwidth]{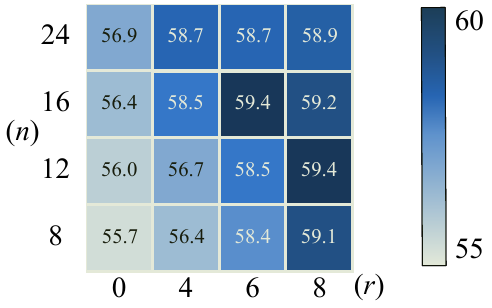}
    \caption{Sensitivity to the Rotation Rank $r$ and Trainable Block Number $n$.}
    \label{fig:n_r}
\end{minipage}
\end{figure}

We present detailed ablation experiments on geometric preservation (\textbf{GP}) and channel-wise scaling (\textbf{CS}) across both benchmarks, shown in Table~\ref{tab:ab_study_dwd} for DWD and Fig.~\ref{fig:ab_study_city} for Cityscapes-C.
The results show that both GP and GIFT (GP combined with CS) improve over the baseline across all target domains (4 for DWD and 15 for Cityscapes-C). Notably, larger gains on Cityscapes-C highlight their effectiveness under more challenging domain shifts.
We also notice that CS further improves performance particularly when built upon a geometrically stable representation, underscoring the importance of preserving stable geometric representations for downstream task-specific adaptation.

\paragraph{Trainable Parameter Efficiency.}

To evaluate the training efficiency of GIFT, we compare the efficiency-effectiveness trade-offs of different methods built upon DINOv2-Large. As shown in Fig.~\ref{fig:param_efficiency}, a fully frozen DINOv2-Large~\cite{dinov2} backbone achieves 54.1\% mPC with 0.0M trainable parameters, whereas GIFT improves mPC by 5.3\% with only 0.52M parameters, demonstrating a superior performance-efficiency balance. Compared to other parameter-efficient fine-tuning strategies, including LoRA~\cite{lora} (2.76M, 54.9\% mPC), Rein~\cite{Rein} (2.99M, 55.5\% mPC), AdaptFormer~\cite{adaptformer} (3.17M, 54.2\% mPC), and VPT~\cite{VPT} (3.69M, 55.5\% mPC), GIFT achieves higher performance while using 6-7$\times$ fewer parameters. Notably, it surpasses the SOTA method Cauvis~\cite{Cauvis} by +1.9\% mPC while reducing trainable parameters by nearly 90\%. These results highlight that GIFT not only enhances robustness but also maintains exceptional parameter efficiency. 

\paragraph{Sensitivity to the Rotation Rank $r$ and Trainable Block Number $n$.}

We investigate the impact of the rotation rank $r$ and the number of trainable blocks $n$ on DWD dataset.\footnote{No target-domain data or labels were used. All configurations were predefined and trained on source-domain data.}
Fig.~\ref{fig:n_r} reports the mPC results across the four adverse weather conditions. We vary $r$ from 0 to 8, where $r=0$ indicates that no rotation matrix is applied. For $n$, we evaluate a wide range from 8 to 24. We observe that the performance generally improves as $r$ increases. However, the gains gradually saturate as $r$ approaches 8, suggesting that a moderately small $r$ is sufficient to capture most of the geometric invariance. We also observe that the performance improves as $n$ increases, reaching the best results when $n$ is between 16 and 24. Based on these observations, we adopt $r=6$ and $n=16$ as the default settings in all main experiments, achieving a favorable balance between performance and computational cost.

\begin{figure}[t]
\centering
\includegraphics[width=0.95\textwidth]{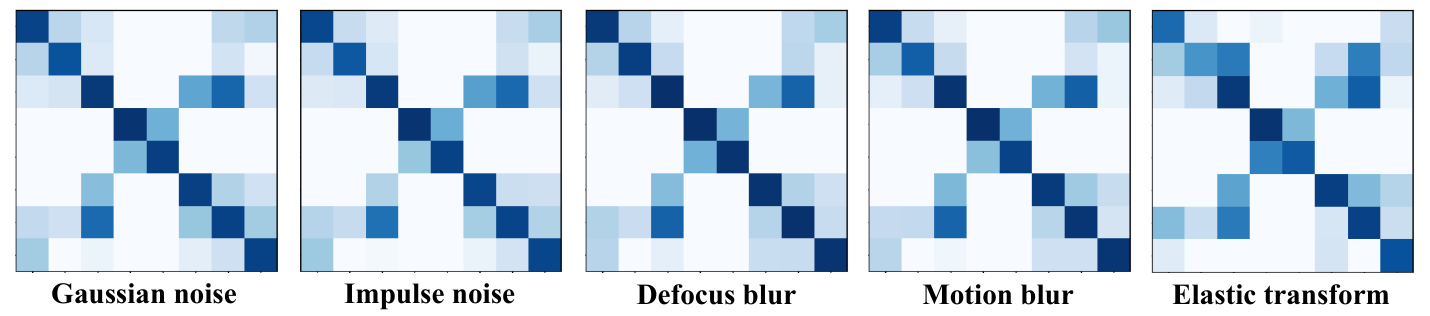}
\caption{Feature correlation analysis on Cityscapes~\cite{cordts2016cityscapes} (Clean domain) and Cityscapes-C~\cite{Cityscapes-C} (Corrupted domain) datasets.
Darker colors indicate higher cosine similarity between class features.}
\label{fig_city_corruption_supp} 
\end{figure}

\paragraph{Feature Correlation Analysis.}

Following OA-DG~\cite{OA-DG}, we measure the feature correlation between the clean and corrupted domains on the Cityscapes~\cite{cordts2016cityscapes} and Cityscapes-C~\cite{Cityscapes-C} to analyze the stability of learned representations under domain corruption. Please refer to the Appendix~\ref{sec:Appendix_Implementation} for more details. We visualize the correlation matrices as heatmaps using five corruption types in Cityscapes-C as examples: `Gaussian noise', `Impulse noise', `Defocus blur', `Motion blur' and `Elastic transform'. As shown in Fig.~\ref{fig_city_corruption_supp}, the correlation matrices exhibit a strong diagonal structure across all corruption types. This indicates that the class-wise representations learned by the model remain largely consistent under various corruption shifts, suggesting that our method could preserve geometric structure of pre-trained representations under domain perturbations.


\begin{table}[t]
\centering
\begin{minipage}[b]{0.5\textwidth}
\centering
\scriptsize
\caption{Comparison of Training Time (s/iter) and Inference Throughput (images/s)}
\label{tab:throughput}
\begin{tabular}{l|c c | c}
\toprule
\textbf{Methods} & Training Time $\downarrow$  & Inf. Throughput $\uparrow$  & \textbf{mPC} \\
\midrule
LoRA~\cite{lora} & \textbf{0.72} & \textbf{6.31} & 54.9 \\
Cauvis~\cite{Cauvis} &1.93  &2.78 &\underline{57.5} \\
GIFT (Ours) & \underline{0.83}  &\underline{5.21}  & \textbf{59.4}\\
\bottomrule
\end{tabular}
\end{minipage}
\hfill
\begin{minipage}[b]{0.48\textwidth}
\centering
\scriptsize
\caption{Comparison of different update schemes for $\bm{W}$ on Cityscapes-C.
}
\label{tab:update_schemes}
\begin{tabular}{l|c c|l l}
\toprule
\textbf{Schemes} & Res. & $\bm{g}$ & Source & Target \\
\midrule
$\bm{W} \rightarrow \bm{W}_{\text{rot}}$ &- & - & 55.8 & 34.2 \\
$\bm{W} \rightarrow \Delta \bm{W}$ & \cmark &- & \underline{57.3}{\tiny(+1.5$\uparrow$)} & \underline{36.5}{\tiny(+2.3$\uparrow$)} \\
$\bm{W} \rightarrow \Delta \bm{W}_{\text{mod}}$ & \cmark & \cmark & \textbf{58.2}{\tiny(+0.9$\uparrow$)} & \textbf{37.6}{\tiny(+1.1$\uparrow$)} \\
\bottomrule
\end{tabular}
\end{minipage}
\end{table}

\paragraph{Comparison of Training Time and Inference Throughput.}

Table \ref{tab:throughput} compares training efficiency and inference speed across different methods. LoRA achieves the fastest training time (0.72 s/iter) and highest inference throughput (6.31 images/s), but its performance is lower than the others. Cauvis improves accuracy (57.5\% in mPC) but at the cost of significantly slower training and inference. In contrast, our method strikes a better balance, achieving competitive efficiency (0.83 s/iter, 5.21 images/s) while obtaining the best performance (59.4\% in mPC).

\paragraph{Comparison of Different Update Schemes.}

In Eq.~(\ref{eq:residual_update}), we employ a residual update mechanism and utilize a learnable gate vector $\bm{g}$ to ensure stable optimization. We analyze the impact of different update mechanisms of the weight $\bm{W}$ on the Cityscapes-C~\cite{Cityscapes-C}.
As shown in Table~\ref{tab:update_schemes}, compared with directly introducing a learnable rotation matrix ($\bm{W} \rightarrow \bm{W}_{\text{rot}}$), applying only a residual update mechanism ($\bm{W} \rightarrow \Delta \bm{W}$) yields an improvement of +1.5\% mAP and +2.3\% mPC, respectively. Moreover, introducing the gate $\bm{g}$ could further improve the performance in the source (from 57.3\% mAP to 58.2\% mAP, +0.9\%) and the target (from 36.5\% mPC to 37.6\% mPC, +1.1\%) domain.

\paragraph{Comparison with Other PEFT Methods.}

Table \ref{tab:different_PEFT} presents a comparison of our method against various PEFT strategies on the DINOv2-Large backbone. Compared with simple freezing (54.1\%), all PEFT methods yield performance improvements, confirming the necessity of PEFT for S-DGOD. Importantly, our method consistently achieves the best performance across all conditions and obtains the highest overall mPC of 59.4\%. For instance, under the most challenging `Night-Rainy' (NR.) conditions, our method reaches 49.9\% in mAP, showing a substantial improvement over LoRA~\cite{lora} (44.7\%, +5.2\%). Furthermore, our method surpasses SOTA method Cauvis~\cite{Cauvis} by +1.9\% in mPC, achieving robust improvements under all four conditions. 


\subsection{Visualization Results}
\label{sec:visual_results}

\begin{figure}[!t]
\centering
\includegraphics[width=0.98\textwidth]{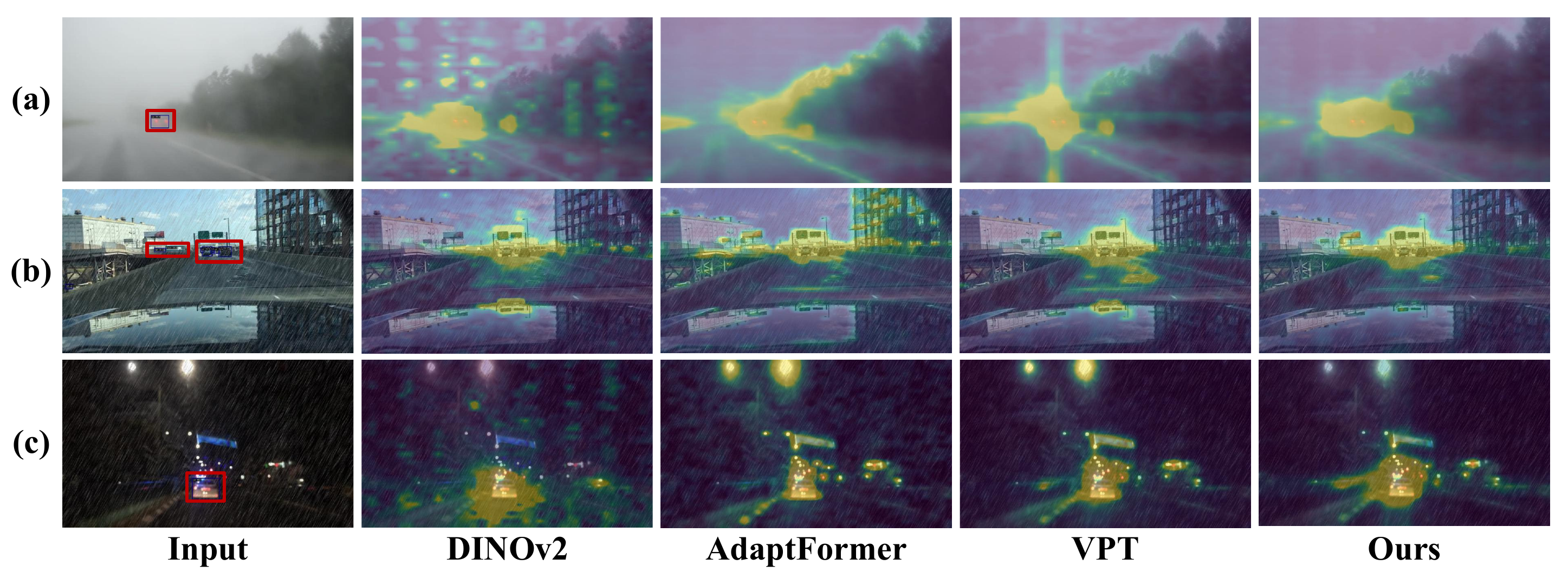}
\caption{Visualization of attention maps for the DINOv2~\cite{dinov2}, conventional PEFT methods AdaptFormer~\cite{adaptformer}, VPT~\cite{VPT}, and our geometric-preserving (GP) fine-tuning strategy on the DWD~\cite{CDSD-DWD}.}
\label{fig_vis_attn}
\end{figure}

\paragraph{Visualization of Attention Maps.}
To further verify the importance of geometric structure of pre-trained representations, we visualize the attention maps of the pre-trained DINOv2~\cite{dinov2}, conventional PEFT methods (AdaptFormer~\cite{adaptformer} and VPT~\cite{VPT}), and our method on the DWD~\cite{CDSD-DWD}. From Fig.~\ref{fig_vis_attn} we can see that the attention maps of the pre-trained DINOv2 contain many artifacts~\cite{darcet2023vision-registers}. Conventional PEFT methods AdaptFormer~\cite{adaptformer} and VPT~\cite{VPT} fail to properly focus on foreground objects, as unconstrained fine-tuning may disrupt the geometric structure of the pre-trained representations. Notably, the attention maps of ours highlight the target regions more clearly, demonstrating that preserving the geometric structure of pre-trained representations is important for domain generalization.

\begin{figure}[t]
\centering
\includegraphics[width=0.98\textwidth]{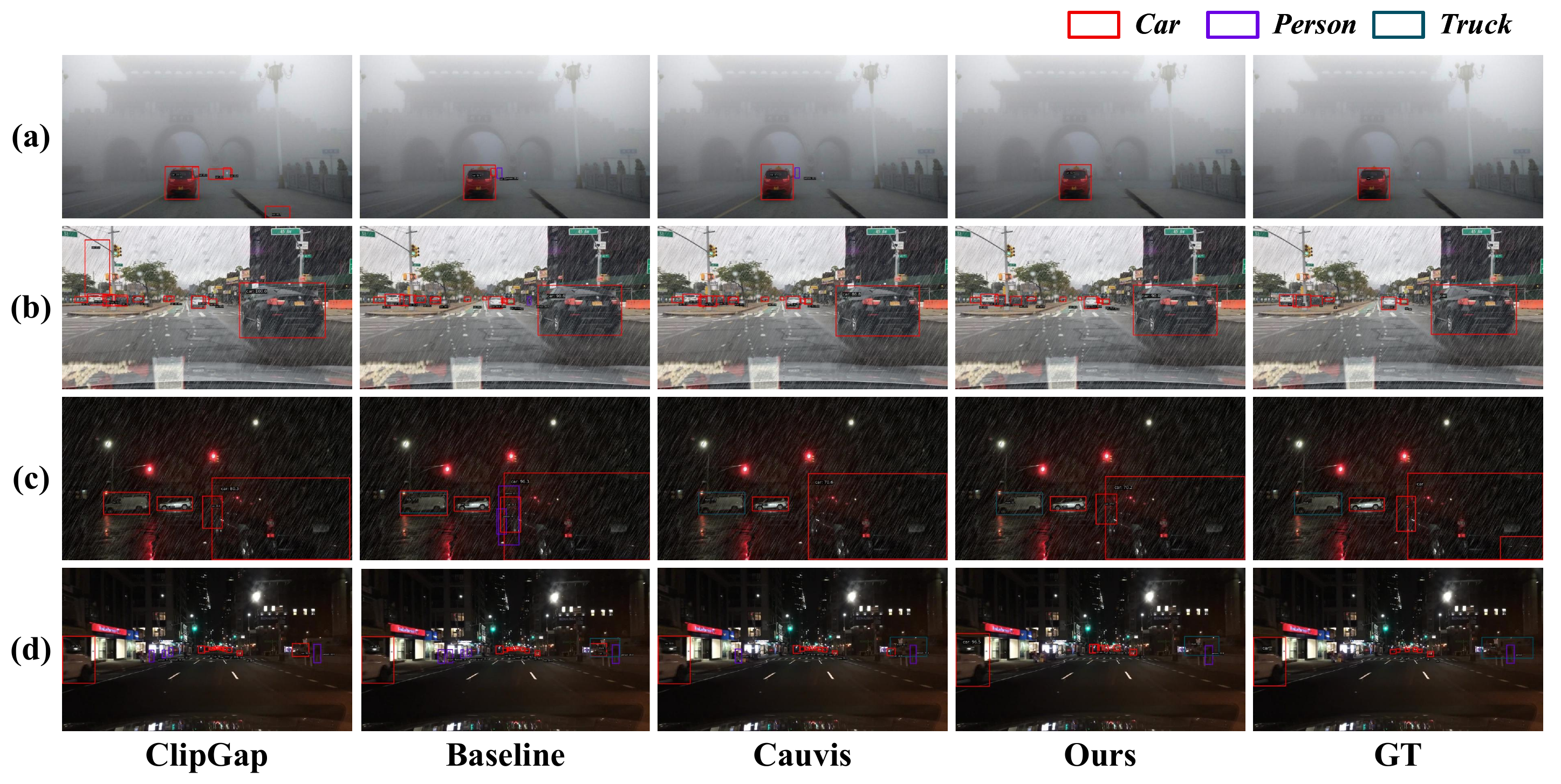}
\caption{Visualization of detection results on the DWD~\cite{CDSD-DWD}. (a) to (d) correspond to the four weather conditions of the DWD target domain: DF., DR., NR., and NS. GIFT is compared with the traditional S-DGOD method ClipGap~\cite{ClipGap}, Baseline (Frozen DINOv2~\cite{dinov2} with Faster R-CNN~\cite{fasterrcnn}) and the SOTA method Cauvis~\cite{Cauvis}. Best seen on computer, in color and zoomed in.}
\label{fig_vis_det} 
\end{figure}

\paragraph{Visualization of Detection Results.}

To show the effectiveness of our method, we provide some visualization results on DWD~\cite{CDSD-DWD}. As shown in Fig.~\ref{fig_vis_det}, existing methods exhibit various detection errors. For example, in Fig.~\ref{fig_vis_det} (d), ClipGap~\cite{ClipGap} fails to detect the truck on the right. Some methods produce false detections of non-existent objects. For example, the baseline detects a non-existent person under daytime-foggy, night-rainy, and night-sunny conditions (see Fig.~\ref{fig_vis_det}(a), (c), (d)). In addition, the SOTA method Cauvis~\cite{Cauvis} suffers from the same issue (Fig.~\ref{fig_vis_det}). Furthermore, other methods misclassify objects, such as in Fig.~\ref{fig_vis_det}(c), where the baseline mis-classifies a truck as a car. In contrast, our method achieves reliable detection performance across all four weather conditions, even in the most challenging night-rainy (NR.) scenario. More results can be found in the Appendix~\ref{Appendix_visual_det}.


\section{Conclusion}

In this work, we revisit Single-Domain Generalized Object Detection (S-DGOD) from the perspective of representation geometry in Vision Foundation Models. We show that unconstrained fine-tuning disrupts pre-trained geometric structures, undermining domain-invariant generalization. To address this, we propose Geometric-Invariance Fine-Tuning (GIFT), a simple yet principled method that preserves representation geometry while enabling effective task adaptation. Experiments on two S-DGOD benchmarks demonstrate that GIFT consistently surpasses state-of-the-art methods, achieving significant performance gains with fewer than 1\% additional trainable parameters. These results highlight the importance of preserving pre-trained representation geometry for robust domain generalization and suggest a promising direction for extending GIFT to broader vision tasks.

\section{Acknowledgments}
This work was conducted during the author's research visit to the Computer Vision Center (CVC). The author sincerely thanks Yaoyue Zheng, Bogdan Raducanu, Joost van de Weijer and many other colleagues in the Learning and Machine Perception (LAMP) team at CVC for their valuable suggestions and discussions.

This work was supported by Grants PID2025-175937NB-I00 and AIA2025-163919-C52 funded by MICIU/AEI/10.13039/501100011033, and by the Generalitat de Catalunya CERCA Program. Yongqiang Zhang acknowledges the National Natural Science Foundation of China (No. 62666051), the Inner Mongolia Natural Science Foundation for Distinguished Young Scholars
(No. 2025JQ009), and the Inner Mongolia Talent Development Project for Outstanding Young Talents. Yaoyue Zheng acknowledges the China Scholarship Council (CSC) No.202406280387.

\bibliographystyle{plain}
\bibliography{neurips_2026}


\appendix

\clearpage

\section*{Appendix}




\section{Additional Ablation Studies and Analysis}
\label{sec:additional_ab}

\subsection{Position of the HR Matrices within the Transformer Blocks.}
In our proposed method, we introduce a learnable rotation matrix $\bm{R}$, defined as the product of $r$ Householder Reflection (HR) matrices, and integrate it into the frozen Transformer blocks. We now investigate the effect of the insertion position of the HR matrices within the Transformer blocks on the S-DGOD task. Specifically, we conduct experiments on the QKV projection layer ($QKV$), the output projection layer ($Output$) and the feed-forward network (MLP layer, $FFN$) within the DINOv2 Transformer blocks. 

As shown in Table~\ref{tab:ab_qkv}, inserting HR matrices only into the $QKV$ projection yields a substantial improvement over the baseline, increasing the mPC from 54.1\% to 58.9\% (first and second rows). This configuration consistently improves performance across all domains, indicating that enhancing the $QKV$ projections strengthens the robustness of attention representations.

Further inserting HR matrices into both the $QKV$ and $Output$ projection layers leads to the best overall performance, achieving the highest mPC of 59.4\% and the best mAP (62.7\%).  In contrast, adding HR matrices to the $FFN$ layer does not bring further gains and slightly degrades performance (mPC drops to 58.7\%). This observation indicates that robustness improvements mainly originate from the attention layers rather than the feed-forward network. Consequently, we adopt the $QKV$ + $Output$ configuration as our default setting.

\begin{table}[htbp]
\centering
\scriptsize
\caption{Position of the HR matrices within the Transformer blocks. First row is Baseline. $QKV$, $Output$, and $FFN$ denote QKV projection layer, output projection layer and feed-forward network (MLP layer), respectively.}
\label{tab:ab_qkv}
\begin{tabular}{ccc|cccccc|c}
\toprule
$QKV$ &$Output$ &$FFN$ & DS. & DF. & DR. & NR. & NS. & Avg. & \textbf{mPC} \\
\midrule
- &- & - & 71.2 & 53.5 & 60.8 & 42.6 & 59.5 & 57.5 & 54.1\\
\cmark &- &-  & \underline{75.1} &\underline{57.5} &65.5 &\textbf{50.1} &62.3 &\underline{62.1} & \underline{58.9}\\
\cmark &\cmark &- & \textbf{76.0} & \textbf{58.3} & \textbf{65.9} & \underline{49.9} & \textbf{63.3} & \textbf{62.7} & \textbf{59.4} \\
\cmark &\cmark &\cmark &75.0 &\underline{57.5} &\underline{65.6} &49.4 &\underline{62.4} &62.0 & 58.7 \\
\bottomrule
\end{tabular}
\end{table}

\subsection{Performance on Domain Generalized Semantic Segmentation.}

We additionally evaluate GIFT on the domain generalized semantic segmentation (DGSS) task. We use Rein~\cite{Rein} as baseline and following previous methods~\cite{Rein, bi2024learning, yun2025soma, zhang2025causal}, we compare our method with existing DGSS methods on two setups: (1) Cityscapes (\textit{C.})~\cite{cordts2016cityscapes} $\rightarrow$ BDD100K (\textit{B.})~\cite{yu2020bdd100k}, Mapillary (\textit{M.})~\cite{neuhold2017mapillary} and (2) GTA5 (\textit{G.})~\cite{richter2016playing} $\rightarrow$ (\textit{C.}), (\textit{B.}), (\textit{M.}). 
As shown in Table~\ref{tab:dgss}, our method achieves the best results in 3 out of 5 cases and the second-best results in 1 case compared to VFM-based methods. These results demonstrate that the proposed geometry-preserving fine-tuning strategy generalizes beyond object detection and is also effective for domain generalized semantic segmentation.

\begin{table}[htbp]
\centering
\scriptsize
\caption{DGSS performance of our proposed GIFT.  (\textit{C.}), (\textit{B.}), (\textit{M.}), and (\textit{G.}) denote Cityscapes, BDD100K, Mapillary and GTA5 datasets, respectively.}
\label{tab:dgss}
\begin{tabular}{l|cc| ccc}
\toprule
\textbf{Methods} &\textit{C.} $\rightarrow $ \textit{B.} &\textit{C.} $\rightarrow $ \textit{M.} & \textit{G.} $\rightarrow $ \textit{C.} & \textit{G.} $\rightarrow $ \textit{B.} & \textit{G.} $\rightarrow $ \textit{M.} \\
\midrule
Rein~\cite{Rein} &63.54 & 74.03 & 66.40 & 60.40 & 66.10\\
FADA~\cite{bi2024learning} &65.12 & 75.86 & 68.23 & \underline{61.94} & 68.09\\
SoMA~\cite{yun2025soma} & \underline{67.02}	&\underline{76.45}	&\textbf{71.82}	&61.31	&\textbf{71.67}\\
Causal-Tune~\cite{zhang2025causal} & 66.28	& 76.05	& 66.22	& 61.80	& \underline{68.21}\\
Ours & \textbf{67.42} & \textbf{76.71} & \underline{68.60} & \textbf{62.01} &68.08\\
\bottomrule
\end{tabular}
\end{table}

\begin{table}[t]
\centering
\scriptsize
\caption{Performance of different VFMs with a DETR-based~\cite{dino-detr-detector} detection head. “Avg.” indicates the mean mAP across the five conditions.}
\label{tab:different_VFM}
\begin{tabular}{ l l| c  c  c c c l | l}
\toprule
\textbf{VFMs} & Fine-tune Method & DS. & DF. & DR. & NR. & NS. & Avg. & \textbf{mPC}\\
\midrule
EVA02-Large~\cite{eva01, eva02} & Freeze & 63.0 & 45.2 & 48.6 & 27.3 & 38.8 & 44.6 & 40.0 \\
& + Rein~\cite{Rein} & 68.3 & 49.2 & 54.8 & 32.2 & \underline{48.1} & 50.5{\tiny (+5.9$\uparrow$)} 
& 46.1{\tiny(+6.1$\uparrow$)} \\
& + Cauvis~\cite{Cauvis} & \underline{69.7} & \underline{50.2} & \underline{57.6} & \underline{34.2} & \underline{48.1} & \underline{52.0}{\tiny (+7.4$\uparrow$)} & \underline{47.5}{\tiny (+7.5$\uparrow$)}\\
& + Ours 
&\textbf{71.6} 
&\textbf{51.3} &\textbf{59.2} &\textbf{35.9} &\textbf{49.2} &\textbf{53.4}{\tiny (+8.8$\uparrow$)} &\textbf{48.9}{\tiny (+8.9$\uparrow$)} \\
\midrule
SAM-Huge~\cite{SAM1} & Freeze & 69.7 & 50.5 & 52.5 & 28.8 & 52.9 & 50.9 & 46.2\\
& + Rein~\cite{Rein} & 70.0 & 51.9 & 54.0 & 30.9 & \underline{54.4} & 52.2{\tiny (+1.3$\uparrow$)} & 47.8{\tiny (+1.6$\uparrow$)}\\
& + Cauvis~\cite{Cauvis} & \underline{72.2} & \textbf{53.7} & \underline{55.8} & \underline{31.9} & \textbf{55.7} & \underline{53.9}{\tiny (+3.0$\uparrow$)}
& \underline{49.3}{\tiny (+3.1$\uparrow$)}\\
& + Ours &\textbf{72.8} &\underline{53.6} &\textbf{56.2} &\textbf{33.1} &\textbf{55.7} & \textbf{54.3}{\tiny (+3.4$\uparrow$)} & \textbf{49.7}{\tiny (+3.5$\uparrow$)} \\
\midrule
DINOv2-Large~\cite{dinov2} & Freeze & 71.2 & 53.5 & 60.8 & 42.6 & 59.5 & 57.5 & 54.1\\
& + Rein~\cite{Rein} & 72.8 & 55.0 & 62.4 & 45.2 & 59.4 & 59.0{\tiny (+1.5$\uparrow$)} 
& 55.5{\tiny (+1.4$\uparrow$)}\\
& + Cauvis~\cite{Cauvis} & \underline{73.7} & \underline{56.5} & \underline{64.6} & \underline{47.6} & \underline{61.2} & \underline{60.7}{\tiny (+3.2$\uparrow$)} & \underline{57.5}{\tiny (+3.4$\uparrow$)}\\
& + Ours & \textbf{76.0} & \textbf{58.3} & \textbf{65.9} & \textbf{49.9} & \textbf{63.3}  & \textbf{62.7}{\tiny (+5.2$\uparrow$)} & \textbf{59.4}{\tiny (+5.3$\uparrow$)} \\
\bottomrule
\end{tabular}
\end{table}

\subsection{Performance across Different VFMs.}
\label{appendix:VFMs}

We further validate the proposed method using three different VFMs, \textit{i.e.}, EVA02-Large~\cite{eva01, eva02}, SAM-Huge~\cite{SAM1}, and DINOv2-Large~\cite{dinov2}. As shown in Table~\ref{tab:different_VFM}, our method achieves strong performance across all three VFM backbones, demonstrating strong generalization. Specifically, our method achieves 48.9\% mPC on the EVA02~\cite{eva02} backbone, surpassing frozen EVA02 by a large margin (+8.9\%). Moreover, it brings a +5.3\% improvement on the DINOv2~\cite{dinov2} backbone and achieves a satisfactory performance (49.7\% mPC) on SAM~\cite{SAM1}. In addition, our method still has significant advantages over the current best method Cauvis~\cite{Cauvis}, achieving improvements of +1.4\%, +0.4\%, and +1.9\% on EVA02, SAM and DINOv2 backbones, respectively. These results indicate that our method enhances the performance of different VFMs and provides stable improvements across diverse weather conditions, highlighting its effectiveness for S-DGOD tasks.




\section{More Implementation Details}
\label{sec:Appendix_Implementation}
Our environment is Ubuntu 22.04, Python 3.9.24, CUDA 11.8, PyTorch 2.2.0, and MMDetection 3.3.0. We adopt the data augmentation pipeline provided by MMDetection and use only the basic augmentation strategy from the original detector for fair comparison. 

For DWD~\cite{CDSD-DWD}, following the standard protocol, Daytime-Sunny is treated as the source domain, containing 18205 training images and 8313 evaluation images. The Daytime-Foggy, Dusk-Rainy, Night-Rainy, and Night-Sunny conditions serve as unseen target domains, with 26158, 2494, 3501, and 3775 images, respectively.
For Cityscapes-C~\cite{Cityscapes-C}, we use the clean Cityscapes~\cite{cordts2016cityscapes} training set as the source domain, containing 2975 training images, while the corrupted images serve as unseen target domains for evaluation.

For more details on the Feature Correlation Analysis in Section~\ref{subsec:ab_exp}, we extract object-level features from the penultimate layer of the classification head and group them according to their object categories. Given an input image, the Transformer backbone produces a sequence of patch tokens. We then compute the average cosine similarity between class-wise features from the clean and corrupted domains to construct the feature correlation matrix.

\section{Detailed Experimental Results}
\label{sec:results_DS}

\subsection{Detailed Results on Diverse Weather Dataset.}

Following Cauvis~\cite{Cauvis}, we group the 7 categories in DWD~\cite{CDSD-DWD} into three groups: Heavy Vehicles `$hev.$' (Bus, Truck), Mid-sized Vehicles `$mid.$' (Car, Motor), and Non-motorized Vehicles `$non.$' (Bike, Person, Rider). `$all$' denotes the mAP computed over all categories under the given weather condition.

\begin{table}[t]
\centering
\caption{Detailed results on Diverse Weather Dataset~\cite{CDSD-DWD}, using Faster R-CNN as the detection head. mPC denotes the average performance across 4 different adverse weather conditions. \textbf{Bold} and \underline{underline} results represent best and second-best performance, respectively. \textbf{GP} denotes geometric preservation.}
\label{tab:supp_DWD}
\resizebox{\textwidth}{!}{ 
\begin{tabular}{l | cccc | cccc | cccc | cccc | c}
\toprule

& \multicolumn{4}{c|}{Daytime-Foggy}
& \multicolumn{4}{c|}{Dusk-Rainy} 
& \multicolumn{4}{c|}{Night-Rainy}
& \multicolumn{4}{c|}{Night-Sunny} \\
\cmidrule(lr){2-5} \cmidrule(lr){6-9} \cmidrule(lr){10-13} \cmidrule(lr){14-17}
\textbf{Methods} & \textit{hev.} & \textit{mid.} & \textit{non.} & \textbf{\textit{all}} & \textit{hev.} & \textit{mid.} & \textit{non.} & \textbf{\textit{all}} & \textit{hev.} & \textit{mid.} & \textit{non.} & \textbf{\textit{all}} & \textit{hev.} & \textit{mid.} & \textit{non.} & \textbf{\textit{all}} &\textbf{mPC} \\
\midrule
IN~\cite{IterNorm} & 25.7 & 33.4 & 27.0 & 28.5 & 34.3 & 25.0 & 13.7 & 22.8 & 19.5 & 11.5 & 8.8 & 12.6 & 38.3 & 27.4  & 25.3 & 29.6 & 23.4 \\
SW~\cite{SW} & 27.1 & 34.9 & 33.8 & 32.2 & 37.0 & 30.3 & 16.6 & 26.3 & 20.0 & 13.9 & 9.4 & 13.7 & 39.5 & 33.2  & 29.6 & 33.4 & 26.1 \\
FRCNN~\cite{fasterrcnn} & 30.6 & 37.8 & 32.6 & 33.5 & 34.6 & 32.0 & 21.1 & 28.0 & 18.7 & 17.5 & 9.0 & 14.2 & 43.3 & 33.7 & 32.2 & 35.8 & 27.9\\
CDSD~\cite{CDSD-DWD} & 28.5 & 39.3 & 32.9 & 33.5 & 38.9 & 32.2 & 18.5 & 28.2 & 21.8 & 19.7 & 11.2 & 16.6 & 42.0 & 35.2 & 34.0 & 36.6 & 28.7 \\
ClipGap~\cite{ClipGap} & 30.7 & 46.0 & 38.9 & 38.5 & 40.1 & 38.8 & 22.8 & 32.3 & 25.8 & 22.7 & 11.3 & 18.7 & 40.3 & 38.6 & 33.5 & 36.9 & 31.6\\
G-NAS~\cite{G-nas} & 28.5 & 44.8 & 36.1 & 36.4 & 45.2 & 40.6 & 24.7 & 35.1 & 25.0 & 19.3 & 11.1 & 17.4 & 47.4 & 47.0 & 42.2 & 45.0 & 33.5\\
PDD~\cite{PDDOC} & 31.2 & 45.9 & 39.7 & 39.1 & 41.7 & 40.7 & 23.9 & 33.7 & 24.3 & 23.0 & 13.1 & 19.2 & 41.9 & 40.2 & 35.1 & 38.5 & 32.6\\
UFR~\cite{UFR} & 32.2 & 47.7 & 39.2 & 39.6 & 40.5 & 42.2 & 22.4 & 33.2 & 26.6 & 22.8 & 11.8 & 19.2 & 45.6 & 40.4 & 37.9 & 40.8 & 33.2\\
\midrule
Baseline & \underline{51.0} & 56.0 & 46.8 & 50.6 & 63.9 & 62.6 & 53.4 & 59.0 & 55.9 & 44.2 & 34.6 & 43.4 & 59.4 & 56.2 & 52.3 & 55.4 & 52.1\\
Cauvis~\cite{Cauvis} & 50.5 & 58.0 & 49.6 & 52.3 & 67.0 & 66.8 & 53.9 & 61.3 & 55.2 & 46.1 & 35.3 & 44.1 & 60.9 & 58.7 & 55.7 & 58.3 & 54.1\\
\textbf{GP Only} & \underline{51.0} & \textbf{61.0} & \textbf{53.9} & \textbf{55.1} & \textbf{68.8} & \underline{66.1} & \textbf{56.7} & \underline{62.8}   & \underline{57.7} & \textbf{51.2} & \underline{36.6} & \underline{46.6} & \underline{62.8} & \underline{61.1} & \underline{59.5} & \underline{60.9} & \underline{56.4} \\

\textbf{GIFT} & \textbf{51.7} & \underline{60.7} & \underline{53.3} & \underline{54.9} & \underline{68.7} & \textbf{68.1} & \textbf{59.1} & \textbf{64.4}   & \textbf{59.0} & \underline{50.8} & \textbf{36.8} & \textbf{47.1} & \textbf{62.9} & \textbf{61.7} & \textbf{61.0} & \textbf{61.7} & \textbf{57.0} \\
\bottomrule
\end{tabular}
}
\end{table}


\begin{table}[htbp]
\centering
\scriptsize
\caption{Detailed results on Daytime-Sunny (source domain) of DWD. Baseline is a frozen DINOv2~\cite{dinov2} backbone with a Faster R-CNN~\cite{fasterrcnn} detector. `F' and `D' denote Faster R-CNN and DETR-based~\cite{dino-detr-detector} detector.}
\label{tab:table_DS}
\begin{tabular}{l | c c c c c c c | c}
\toprule
 & \multicolumn{8}{c}{Daytime-Sunny} \\
\cmidrule(lr){2-9}
\textbf{Methods} & Bus & Bike & Car & Motor & Person & Rider & Truck & \textbf{mAP} \\
\midrule
InterNorm~\cite{IterNorm} &58.4 &34.2 &42.2 &44.1 &31.6 &40.8 &55.5 &43.9\\
SW~\cite{SW} & 62.3 &42.9 &53.3 &49.9 &39.2 &46.2 &60.6 &50.6\\
ISW~\cite{ISW} &62.9 &44.6 &53.5 &49.2 &39.9 &48.3 &60.9 &51.3\\
Faster R-CNN~\cite{fasterrcnn} &66.9 &45.9 &69.8 &46.5 &50.6 &49.6 &64.0 &56.2 \\
CDSD~\cite{CDSD-DWD} &68.8 &50.9 &53.9 &56.2 &41.8 &52.4 &68.7 &56.1 \\
ClipGap~\cite{ClipGap} &55.0 &47.8 &67.5 &46.7 &49.4 &46.7 &54.7 &52.5 \\
UFR~\cite{UFR} &66.8 &51.0 &70.6 &55.8 &49.8 &48.5 &67.4 &58.6 \\
Cauvis~\cite{Cauvis} (D) &76.3 &\underline{64.3} &87.7 &\underline{64.2} &\underline{79.2} &72.1 &71.8 &73.7 \\
\midrule
Ours (F) &\underline{77.3} &61.3 &\underline{89.1} &63.8 &77.5 &\underline{76.0} &\underline{72.4} &\underline{73.9} \\
Ours (D) &\textbf{77.6} &\textbf{67.0} &87.5 &\textbf{68.9} &\textbf{80.7} &\textbf{76.3} &\textbf{74.1} &\textbf{76.0} \\
\bottomrule
\end{tabular}
\end{table}

As shown in Table~\ref{tab:supp_DWD}, under diverse weather-induced domain shifts, the proposed method consistently achieves the best performance across four unseen target domains. In particular, GIFT attains the highest mPC of 57.0\%, surpassing the previous SOTA method, Cauvis~\cite{Cauvis}, by 2.9\%, and outperforming the baseline (Frozen DINOv2~\cite{dinov2} with Faster R-CNN~\cite{fasterrcnn}) by 4.9\%. In addition, GIFT demonstrates a significant performance margin over traditional S-DGOD methods such as UFR~\cite{UFR}(+23.8\%), PDD~\cite{PDDOC}(+24.4\%), and G-NAS~\cite{G-nas}(+23.5\%), showing that the proposed VFM-based fine-tuning strategy effectively addresses the main bottleneck in S-DGOD.

Notably, even when only geometric preservation is applied, the model still outperforms previous methods. Across the four unseen domains, it improves over the baseline by 4.5\%, 3.8\%, 3.2\%, and 5.5\%, and surpasses the SOTA method Cauvis~\cite{Cauvis} by 2.8\%, 1.5\%, 2.5\%, and 2.6\%, highlighting the effectiveness of preserving the geometric structure of pre-trained representations.

\subsection{Results on Daytime-Sunny.}
We report the detection results on the source domain of DWD~\cite{CDSD-DWD}. As shown in Table~\ref{tab:table_DS}, our method significantly outperforms traditional S-DGOD methods in terms of mAP. For example, when using the same detector, Faster R-CNN~\cite{fasterrcnn}, our method achieves improvements of +15.3\% over UFR~\cite{UFR} (from 58.6\% to 73.9\%) and +21.4\% over ClipGap~\cite{ClipGap} (52.5\% \textit{vs.} 73.9\%). Surprisingly, our method with Faster R-CNN detector is even superior to Cauvis~\cite{Cauvis}, which uses a DETR-based detector (73.9\% \textit{vs.} 73.7\%). Moreover, when using a DETR-based detector, our method achieves a new state-of-the-art result of 76\% in mAP, surpassing Cauvis by 2.3\%, and achieves the best performance in 6 out of the 7 categories.

\begin{table}[t]
\centering
\scriptsize
\caption{Detailed results on unseen target domain of DWD. `Mot', `Per.', `Rid.', and `Tru.' denote `Motor', `Person', `Rider', and `Truck', respectively. `F' and `D' denote Faster R-CNN~\cite{fasterrcnn} and DETR-based~\cite{dino-detr-detector} detector.}
\label{tab:table_target}
\resizebox{\textwidth}{!}{
\begin{tabular}{l | c c c c c c c | c || c c c c c c c | c}
\toprule
 & \multicolumn{8}{c||}{Daytime-Foggy} & \multicolumn{8}{c}{Dusk-Rainy}\\
\cmidrule(lr){2-9} \cmidrule(lr){10-17}
\textbf{Methods} & Bus & Bike & Car & Mot. & Per. & Rid. & Tru. & \textbf{mAP} & Bus & Bike & Car & Mot. & Per. & Rid. & Tru. & \textbf{mAP}\\
\midrule
FR~\cite{fasterrcnn} &34.5 &29.6 &49.3 &26.2 &33.0 &35.1 &26.7 &33.5 &34.2 &21.8 &47.9 &16.0 &22.9 &18.5 &34.9 &28.0 \\
CDSD~\cite{CDSD-DWD} &32.9 &28.0 &48.8 &29.8 &32.5 &38.2 &24.1 &33.5 &37.1 &19.6 &50.9 &13.4 &19.7 &16.3 &40.7 &28.2 \\
ClipGap~\cite{ClipGap} &36.2 &34.2 &57.9 &34.0 &38.7 &43.8 &25.1 &38.5 &37.8 &22.8 &60.7 &16.8 &26.8 &18.7 &42.4 &32.3 \\
G-NAS~\cite{G-nas} &32.4 &31.2 &57.7 &31.9 &38.6 &38.5 &24.5 &36.4 &44.6 &22.3 &66.4 &14.7 &32.1 &19.6 &45.8 &35.1 \\
PDD~\cite{PDDOC} &36.1 &34.5 &58.4 &33.3 &40.5 &44.2 &26.2 &39.1 &39.4 &25.2 &60.9 &20.4 &29.9 &16.5 &43.9 &33.7 \\
UFR~\cite{UFR} &36.9 &35.8 &61.7 &33.7 &39.5 &42.2 &27.5 &39.6 &37.1 &21.8 &67.9 &16.4 &27.4 &17.9 &43.9 &33.2 \\
Cauvis~\cite{Cauvis} (D) &50.7 &\underline{43.8} &73.8 &\underline{50.5} &\underline{67.4} &\underline{61.9} &47.5 &\underline{56.5} &67.2 &\underline{54.8} &85.6 &\underline{52.0} &\underline{65.5} &\underline{57.2} &\underline{70.1} &\underline{64.6} \\
\midrule
Ours (F) &\underline{54.2} &38.7 &\underline{73.9} &47.4 &61.9 &59.3 &\textbf{49.1} &54.9  &\underline{67.3} &51.8 &\textbf{87.1} &49.0 &65.4 &\textbf{60.2} &70.0 &64.4\\
Ours (D) &\textbf{55.0} &\textbf{46.1} &\textbf{74.1} &\textbf{51.7} &\textbf{68.8} &\textbf{64.3} &\underline{48.0} &\textbf{58.3} &\textbf{68.5} &\textbf{55.2} &\underline{85.7} &\textbf{57.9} &\textbf{66.8} &56.4 &\textbf{71.0} &\textbf{65.9} \\
\midrule
\midrule
 & \multicolumn{8}{c||}{Night-Rainy} & \multicolumn{8}{c}{Night-Sunny}\\
\cmidrule(lr){2-9} \cmidrule(lr){10-17}
\textbf{Methods} & Bus & Bike & Car & Mot. & Per. & Rid. & Tru. & \textbf{mAP} & Bus & Bike & Car & Mot. & Per. & Rid. & Tru. & \textbf{mAP}\\
\midrule
FR~\cite{fasterrcnn} &21.3 &7.7 &28.8 &6.1 &8.9 &10.3 &16.0 &14.2 &43.5 &31.2 &49.8 &17.5 &36.3 &29.2 &43.1 &35.8 \\
CDSD~\cite{CDSD-DWD} &24.4 &11.6 &29.5 &9.8 &10.5 &11.4 &19.2 &16.6 &40.6 &35.1 &50.7 &19.7 &34.7 &32.1 &43.4 &36.6 \\
ClipGap~\cite{ClipGap} &28.6 &12.1 &36.1 &9.2 &12.3 &9.6 &22.9 &18.7 &37.7 &34.3 &58.0 &19.2 &37.6 &28.5 &42.9 &36.9 \\
G-NAS~\cite{G-nas} &28.6 &9.8 &38.4 &0.1 &13.8 &9.8 &21.4 &17.4 &46.9 &40.5 &67.5 &26.5 &50.7 &35.4 &47.8 &45.0 \\
PDD~\cite{PDDOC} &25.6 &12.1 &35.8 &10.1 &14.2 &12.9 &22.9 &19.2 &40.9 &35.0 &59.0 &21.3 &40.4 &29.9 &42.9 &38.5 \\
UFR~\cite{UFR} &29.9 &11.8 &36.1 &9.4 &13.1 &10.5 &23.3 &19.2 &43.6 &38.1 &66.1 &14.7 &49.1 &26.4 &47.5 &40.8  \\
Cauvis~\cite{Cauvis} (D) &60.8 &\underline{32.4} &69.5 &24.0 &\underline{49.9} &\textbf{39.5} &\underline{57.2} &\underline{47.6} &\underline{62.8} &\underline{55.1} &79.5 &39.3 &\underline{70.4} &57.4 &\underline{64.2} &61.2 \\
\midrule
Ours (F) &\underline{61.5} &29.5 &\textbf{73.0} &\textbf{28.6} &48.8 &32.2 &56.5 &47.1 &\underline{62.8} &53.9 &\textbf{83.5} &\underline{39.8} &66.8 &\textbf{62.3} &63.0 &\underline{61.7} \\
Ours (D) &\textbf{64.7} &\textbf{38.5} &\underline{70.9} &\underline{27.1} &\textbf{50.0} &\underline{37.3} &\textbf{61.0} &\textbf{49.9} &\textbf{63.9} &\textbf{59.3} &\underline{80.5} &\textbf{40.5} &\textbf{72.6} &\underline{60.6} &\textbf{65.5} &\textbf{63.3} \\
\bottomrule
\end{tabular}
}
\end{table}

\subsection{Results on Unseen Target Domain.}

Table~\ref{tab:table_target} presents a comprehensive evaluation of our proposed method compared with several S-DGOD methods across four  unseen target domains in DWD~\cite{CDSD-DWD}.

\textbf{Daytime-Foggy.} Our DETR-based model (Ours (D)) achieves the best overall performance with 58.3\% mAP, surpassing the previous state-of-the-art method Cauvis~\cite{Cauvis} by +1.8\% mAP. Specifically, our method consistently improves most object categories, including Bus (55.0\% \textit{vs.} 50.7\%), Bike (46.1\% \textit{vs.} 43.8\%), Motor (51.7\% \textit{vs.} 50.5\%), Person (68.8\% \textit{vs.} 67.4\%), and Rider (64.3\% \textit{vs.} 61.9\%). Notably, our Faster R-CNN-based version (Ours (F)) also performs competitively, particularly in the Truck category, where it achieves the highest accuracy of 49.1\%. Despite the reduced visibility caused by fog, our method still demonstrates strong robustness.

\textbf{Dusk-Rainy.} Our method achieves 65.9\% mAP, outperforming Cauvis~\cite{Cauvis} by +1.3\% mAP. It is worth noting that Ours (F) exhibits exceptional performance in the Car category, reaching 87.1\% mAP, and the Rider category at 60.2\%. These results suggest that our framework effectively preserves geometric structure even when visual clarity is compromised by rain artifacts and low-light conditions during dusk.

\textbf{Night-Rainy.} The Night-Rainy domain is the most challenging scenario due to extremely low illumination and rain artifacts. Although all models experience performance degradation, Ours (D) maintains a clear lead with 49.9\% mAP. In particular, the improvement on Bike (from 32.4\% to 38.5\%, +6.1\% mAP) demonstrates the effectiveness of our method in detecting small objects under severe visual degradation.

\textbf{Night-Sunny.} Our models continue to dominate the results under Night-Sunny conditions. Ours (D) achieves the highest mAP of 63.3\%, showing superior performance in the Person (72.6\%) and Bike (59.3\%) categories. These results demonstrate that our method effectively handles object appearance variations caused by strong illumination contrast in nighttime scenes.

Overall, the consistent improvements across all  domains demonstrate that our method effectively enhances cross-domain robustness and generalization under diverse adverse weather and illumination conditions.

\begin{figure}[htbp]
\centering
\includegraphics[width=0.98\textwidth]{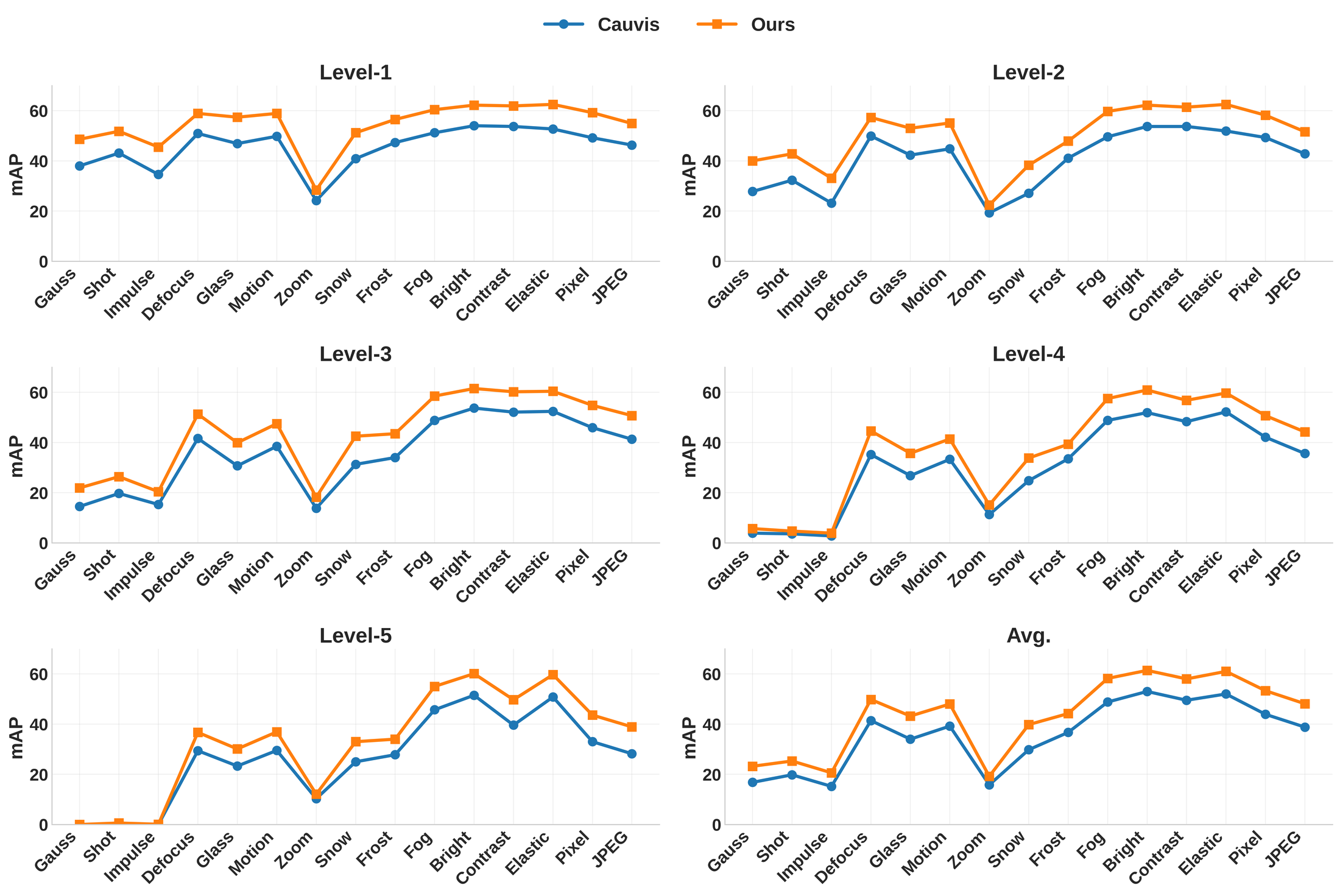}
\caption{Comparison between our method and Cauvis~\cite{Cauvis} on Cityscapes-C from Level-1 to Level-5. `Avg.' denotes the average mAP across the 5 corruption levels.}
\label{fig_vis_corruption_level_supp} 
\end{figure}

\begin{table}[!t]
\centering
\scriptsize
\caption{Quantitative results on Cityscapes-C from Level-1 to Level-5. `Avg.' denotes the average mAP across the 5 corruption levels.}
\label{tab:city_1-5}
\resizebox{\textwidth}{!}{
\begin{tabular}{l|cccccc||cccccc}
\toprule
 &\multicolumn{6}{c||}{GIFT (Ours)} & \multicolumn{6}{c}{Cauvis~\cite{Cauvis}}\\
\cmidrule(lr){2-7} \cmidrule(lr){8-13} 
\textbf{Types} & Level-1 & Level-2 & Level-3 & Level-4 & Level-5 &Avg. & Level-1 & Level-2 & Level-3 & Level-4 & Level-5 &Avg.\\
\midrule
Source &- &- &- &- &- &62.7 &- &- &- &- &- &54.6 \\
\midrule
Gauss &48.6 &40.0 &21.9 &5.7 &0.0 &23.2 &38.0 &27.8 &14.5 &3.9 &0.0  &16.8 \\
Shot &51.2 &42.8 &26.4 &4.7 &0.6 &25.1 &43.1 &32.3 &19.7 &3.6 &0.1 &19.8 \\
Impulse &45.5 &33.1 &20.4 &3.9 &0.1 &20.6  &34.6 &23.2 &15.3 &2.8 &0.0 &15.2 \\
Defocus &58.9 &57.3 &51.3 &44.6 &36.7 &49.8 &50.9 &49.9 &41.6 &35.2 &29.4 &41.4 \\
Glass &57.4 &53.0 &39.9 &35.7 &30.1 &43.2 &46.9 &42.3 &30.7 &26.8 &23.3 &34.0 \\
Motion &58.9 &55.1 &47.5 &41.4 &36.9 &48.0 &49.8 &44.8 &38.5 &33.3 &29.5 &39.2 \\
Zoom &28.4 &22.4 &18.2 &15.1 &12.1 &19.2 &24.2 &19.3 &13.8 &11.3 &10.3 &15.8 \\
Snow &51.2 &38.3 &42.5 &33.8 &33.0 &39.8 &40.9 &27.1 &31.3 &24.8 &25.0 &29.8 \\
Frost &56.5 &47.9 &43.5 &39.3 &34.0 &44.2 &47.3 &41.1 &34.0 &33.5 &27.8 &36.7 \\
Fog &60.4 &59.7 &58.5 &57.5 &55.0 &58.2 &51.2 &49.6 &48.8 &48.8 &45.7 &48.8 \\
Bright &62.2 &62.2 &61.5 &60.9 &60.1 &61.4 &54.0 &53.7 &53.7 &51.9 &51.5 &53.0 \\
Contrast &61.9 &61.4 &60.2 &56.8 &49.7 &58.0 &53.7 &53.7 &52.1 &48.3 &39.6 &49.5 \\
Elastic &62.5 &62.5 &60.4 &59.7 &59.7 &61.0 &52.7 &51.9 &52.4 &52.2 &50.8 &52.0 \\
Pixel &59.2 &58.2 &54.8 &50.7 &43.6 &53.3 &49.2 &49.3 &45.9 &42.1 &33.0 &43.9 \\
JPEG &54.9 &51.6 &50.7 &44.2 &38.9 &48.1 &46.3 &42.8 &41.3 &35.6 &28.2 &38.8 \\
\midrule
\textbf{mPC} &- &- &- &- &- &\textbf{43.5} &- &- &- &- &- &\underline{35.6} \\
\bottomrule
\end{tabular}
}
\end{table}

\begin{figure}[!t]
\centering
\includegraphics[width=0.98\textwidth]{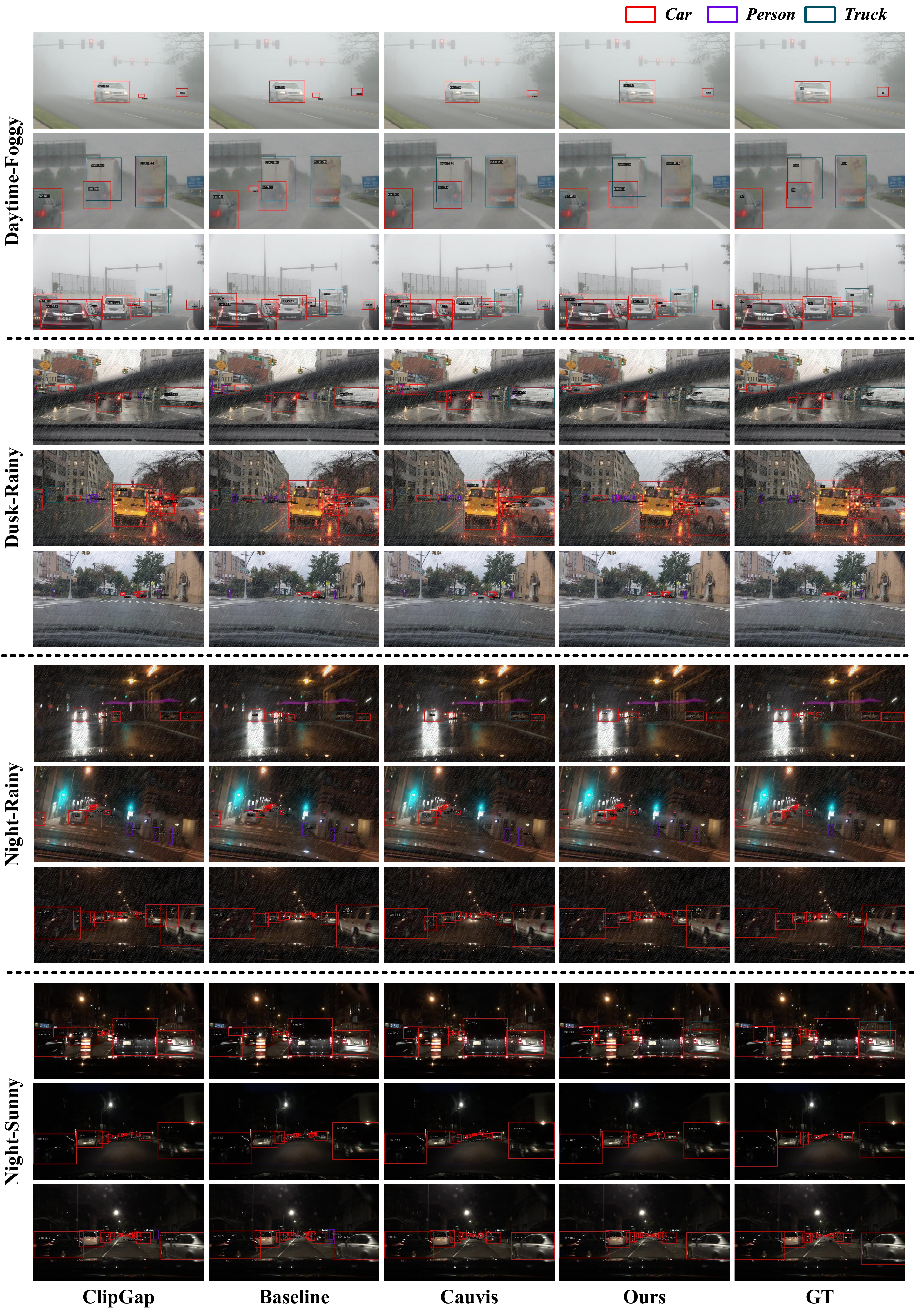}
\caption{More visualization of detection results on the DWD~\cite{CDSD-DWD}. (a) to (d) correspond to the four weather conditions of the DWD target domain: DF., DR., NR., and NS. GIFT is compared with the traditional S-DGOD method ClipGap~\cite{ClipGap}, Baseline (Frozen DINOv2~\cite{dinov2} with Faster R-CNN~\cite{fasterrcnn}) and the SOTA method Cauvis~\cite{Cauvis}. Best seen on computer, in color and zoomed in.}
\label{fig_vis_det_supp} 
\end{figure}

\subsection{Results on Cityscapes-C from Level-1 to Level-5.}

Fig.~\ref{fig_vis_corruption_level_supp} and Table~\ref{tab:city_1-5} present a detailed comparison between our method and Cauvis~\cite{Cauvis} on the Cityscapes-C~\cite{Cityscapes-C} dataset across 5 severity levels. As shown in Fig.~\ref{fig_vis_corruption_level_supp}, our method consistently outperforms Cauvis across almost all corruption types and severity levels. Specifically, our method achieves higher mAP under both Noise-based corruptions (e.g., Gaussian, Shot, and Impulse) and Blur-based corruptions (e.g., Defocus, Glass, Motion, and Zoom), demonstrating stronger robustness to various perturbations.

As the corruption severity increases from Level-1 to Level-5, the performance of both methods gradually decreases. However, our method exhibits a slower degradation trend compared to Cauvis. For instance, under severe corruptions such as Level-5, our method still maintains noticeably higher results across most corruption types. Moreover, as shown in Table~\ref{tab:city_1-5}, our method achieves an excellent performance (43.5\% in mPC) compared with Cauvis (35.6\%), showing an overall improvement of +7.9\%. These results demonstrate that our method substantially improves the robustness of the model against diverse corruptions.

\section{More Visualization of Detection Results}
\label{Appendix_visual_det}

To further demonstrate the robustness of our method, we provide more visualization of detection results across four challenging target domains: Daytime-Foggy, Dusk-Rainy, Night-Rainy, and Night-Sunny in Fig.~\ref{fig_vis_det_supp}.

\textbf{Daytime-Foggy.} In foggy conditions characterized by low contrast and obscured details, our method demonstrates superior capability in identifying distant cars and trucks that are often incorrectly detected by ClipGap~\cite{ClipGap} and the Baseline. Moreover, the Baseline and Cauvis~\cite{Cauvis} often produce inaccurate bounding boxes due to the degraded visibility while our model remains well aligned with the ground truth (GT).

\textbf{Dusk-Rainy.} In this scenario, other models frequently produce incomplete detections or inaccurate localization due to reduced visibility and illumination variations. In contrast, our method accurately detects persons and trucks despite the visual noise caused by rain streaks, indicating improved robustness to simultaneous illumination and weather variations.

\textbf{Night-Rainy.} In this setting, most other methods suffer from severe detection failures. For instance, ClipGap~\cite{ClipGap}, Baseline and Cauvis~\cite{Cauvis} mistakenly detect cars as trucks. In contrast, our model shows remarkable resilience against glare from streetlights and headlights. These results demonstrate that the proposed method can better preserve geometric structure even under extreme nighttime weather conditions.

\textbf{Night-Sunny.} Although this domain does not contain rain or fog, extremely low illumination still poses significant challenges for object detection. Many other methods fail to detect small cars or distant trucks while our predictions are nearly indistinguishable from the GT.

\section{Limitation}
\label{sec:Appendix_limitation}

First, when the pre-trained geometry is not well aligned with the downstream task, the relative improvement may be smaller than that observed in natural RGB driving scenarios, 
such as infrared or remote sensing images that differ substantially from natural-image pretraining data. Second, although GIFT reduces the number of trainable parameters and improves adaptation efficiency, its reliance on large-scale pre-trained VFMs still requires considerable GPU memory. In addition, this work focuses on a fixed category set. Its effectiveness under open-vocabulary domain generalization, where the detector needs to generalize to both unseen domains and novel categories, remains unexplored. Extending the proposed framework to open-vocabulary domain generalization is an important direction for future work.



\end{document}